%% file: main.tex
\pdfoutput=1

\documentclass[11pt]{article}

\input{packages}

\title{Towards Interactive Language Modeling}

\newcommand{\ruimte}{ \ }

  \author{
    \mbox{}$\!\!\!$Maartje ter Hoeve$^{1, *}$\ruimte
    Evgeny Kharitonov$^{2}$\ruimte
    Dieuwke Hupkes$^3$\ruimte
    Emmanuel Dupoux$^3$
    $\!\!\!$\mbox{} \\
    $^1$University of Amsterdam\quad
    $^2$Work done while at Meta AI Labs\\\quad
    $^3$Meta AI Labs\\
    \href{mailto:m.a.terhoeve@uva.nl}{m.a.terhoeve@uva.nl}, 
    \href{mailto:eugene.kharitonov@gmail.com}{eugene.kharitonov@gmail.com}, 
    \href{mailto:dieuwkehupkes@fb.com}{dieuwkehupkes@fb.com},
    \href{mailto:dpx@fb.com}{dpx@fb.com}}

\begin{document}
\maketitle

\blfootnote{*Work done while interning at Meta AI Labs.}

\begin{abstract}

Interaction between caregivers and children plays a critical role in human language acquisition and development. Given this observation, it is remarkable that explicit interaction plays little to no role in artificial language modeling---which also targets the acquisition of human language, yet by artificial models. Moreover, an interactive approach to language modeling has the potential to make language models substantially more versatile and to considerably impact downstream applications. Motivated by these considerations, we pioneer the space of interactive language modeling. First we present a road map in which we detail the steps that need to be taken towards interactive language modeling. We then lead by example and take the first steps on this road map, showing the initial feasibility of our approach. As such, this work aims to be the start of a larger research agenda on interactive language modeling.

\end{abstract}

\input{sections/01_introduction}
\input{sections/02_related_work}

\input{sections/03_general_method}
\input{sections/04_our_approach}

\input{sections/05_experimental_setup}
\input{sections/06_results}
\input{sections/07_implications}
\input{sections/08_conclusion}
\input{sections/09_ethical_impact}

\section*{Acknowledgements}
We thank Léonard Blier, Marco Baroni, Armand Joulin, Douwe Kiela, and Adina Williams for helpful discussions and suggestions.

\bibliographystyle{acl_natbib}
\bibliography{anthology,custom}

\appendix
\input{sections/10_appendix}

\end{document}

%% file: packages.tex
\usepackage{acl}

\usepackage{times}
\usepackage{latexsym}

\usepackage[T1]{fontenc}

\usepackage[utf8]{inputenc}

\usepackage{microtype}

\usepackage{xcolor}
\usepackage{tabularx}
\usepackage{colortbl}
\usepackage{tabularx}
\usepackage{enumerate}
\usepackage[inline]{enumitem}
\usepackage{array}

\usepackage{amsmath, amssymb, amscd, amsthm, amsfonts}
\usepackage{graphicx}
\usepackage{hyperref}
\usepackage{xspace}
\usepackage{xcolor}
\usepackage{array}
\usepackage{booktabs}
\usepackage{caption}
\usepackage{subcaption}

\newcommand\blfootnote[1]{%
  \begingroup
  \renewcommand\thefootnote{}\footnote{#1}%
  \addtocounter{footnote}{-1}%
  \endgroup
}

%% file: sections/01_introduction.tex

\section{Introduction}
\label{sec:introduction}

Interaction between children and more advanced language interlocutors (such as caregivers) plays an important role in many theories and studies on human language acquisition~\cite[e.g.,][]{bruner1985child, clark2018conversation}. For example, although culturally dependent~\cite{shneidman2012language} and with the precise effects still up for discussion~\cite{cristia2019segmentability}, caregivers can communicate with their children in Child Directed Speech. In turn, children can for example experiment with the meaning of words, to elicit a response from their caregivers~\cite{gillis2000kindertaalverwerving}.

\begin{figure}
    \centering
    \includegraphics[width=0.45\textwidth]{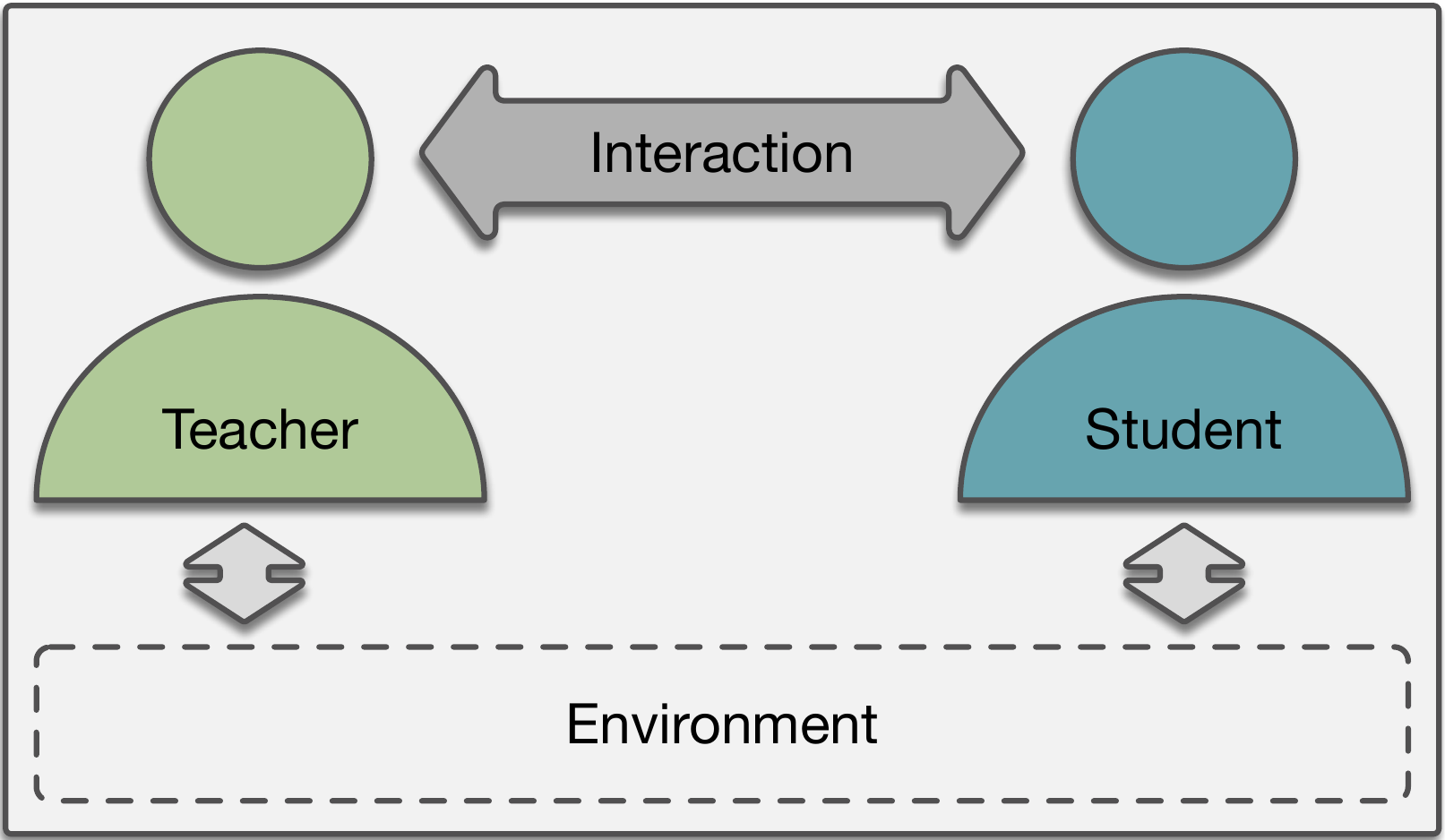}
    \caption{Teacher-Student setup for interactive language modeling.}
    \label{fig:teacher-student-setup-general}
    \vspace{-1em}
\end{figure}

Despite the importance of interaction in human language acquisition, interaction plays little to no role in artificial language modeling. This is remarkable, as language modeling also has the objective to learn human language, albeit with artificial models. Instead, current state-of-the-art language models (LMs) take large amounts of text as input, and are tasked to predict the next or masked words~\cite[e.g.,][]{devlin-etal-2019-bert, brown2020language}. The learning signal only comes from a cross-entropy loss that indicates whether a prediction was correct. Although this setup has shown to be effective, from the perspective of human language acquisition it appears very unnatural. This gives rise to the motivation to investigate other, more natural approaches to language modeling, such as the interactive perspective that we propose in this paper.

Specifically, we structure our proposal according to a teacher-student setup. Figure~\ref{fig:teacher-student-setup-general} depicts a high level overview. In this setup we distinguish four main parts: \textit{the teacher}, whose role is inspired by the caregiver in the human language acquisition, \textit{the student}, who resembles the child, \textit{the interaction} between the teacher and the student and \textit{the environment} that they both share (such as the language that needs to be learned by the student). We motivate and detail our setup further in Section~\ref{sec:interactive_lm}.

An interactive approach to language modeling is not only interesting from the perspective of human language acquisition. Explicitly allowing for interaction also has the potential to make language modeling more efficient and versatile. For example, a teacher can adapt its input to a student based on the specific feedback signals it receives from the student, and a teacher that is fluent in one domain can teach the specifics of that domain to a student trained on another domain, and vice versa. Moreover, an interactive approach to language modeling has the potential to impact downstream applications, for example for foreign language teaching apps where a student can be replaced by a human.

In this paper we pioneer the space of interactive language modeling. Specifically, we contribute:

\begin{enumerate}[leftmargin=*,label=\textbf{C\arabic*},nosep]
\item We define the objective of interactive language modeling;
\item We present a road map that details the steps that need to be taken towards this objective;
\item We take the first steps on this road map, which show the initial feasibility of our approach.
\end{enumerate}

By doing so we aim to start a larger research agenda on interactive language modeling.

%% file: sections/02_related_work.tex

\section{Related Work}

Over the years many different types of learning strategies have been proposed for artificial modeling. Below we describe a number of them that are particularly related to the current work.

\subsection{Interactive Language Learning in NLP}

Recently, a number of studies have focused on interactive language learning. \citet{stein-etal-2021-shapelurn} learn logical semantic representations in an interactive way. \citet{nikolaus-fourtassi-2021-modeling} propose a proof of concept to model perception and production based learning of semantic knowledge acquisition in children. \citet{kiseleva2022interactive, kiseleva2022iglu} take an interactive approach to language \textit{understanding} in a recent NeurIPS challenge. To the best of our knowledge, none of the existing works have focused specifically on language modeling.

\subsection{Curriculum Learning}
Curriculum Learning (CL)~\cite{bengio2009curriculum} is an approach to learning in which data samples are presented in a meaningful order---typically in order of complexity---motivated by the idea that humans learn in a similar way.~\citeauthor{bengio2009curriculum} show the effectiveness of CL on a number of tasks, among which a classical approach to language modeling. More recently, a number of studies have shown the effectiveness of CL for (fine-tuning) LMs~\cite{xu-etal-2020-curriculum, zhang2021reducing}, although other studies have shown that not all intuitive curricula are also effective~\cite{liu2019roberta}. \citet{matiisen2019teacher} propose a teacher-student framework for automatic CL for the addition of decimal numbers and navigation in Minecraft. 

\subsection{Active Learning}
In Active Learning (AL)~\cite{cohn1996active} a learner (the model to be trained) actively selects which data it can most effectively be trained on. That is, where CL is often more associated with choosing a teaching strategy, AL is rather focused on the student side. AL is often used to efficiently label data in a low resource setting~\cite[e.g.,][]{reichart-etal-2008-multi, ein-dor-etal-2020-active}.

\subsection{Continual Learning}
In Continual Learning, or life-long learning, the aim is to train a model in an online fashion, i.e., on a continuous stream of data, whilst avoiding \textit{catastrophic forgetting}~\cite{mccloskey1989catastrophic, french1999catastrophic}. This makes models versatile to an ever changing world. Some recent work has focused on types of Continual Learning for large LMs~\cite[e.g.,][]{lazaridou2021pitfalls, jin-etal-2022-lifelong-pretraining}. We envision interactive language modeling to play an important role in life-long learning in the future.

%% file: sections/03_general_method.tex

\section{A Road Map towards Interactive Language Modeling}
\label{sec:interactive_lm}

In this section we present a general road map towards interactive language modeling. 

\vspace{5pt}
\noindent \textit{Our objective is to build an automated teacher-student loop for language modeling that attains good performance in the student for a fixed (low) number of bits transmitted in the interactions.}
\vspace{5pt}

\noindent We propose a teacher-student loop as this format closely resembles caregiver-child interactions. In Section~\ref{sec:introduction} and Figure~\ref{fig:teacher-student-setup-general} we already introduced a high level overview of this setup and its four main components: (1) \textit{the teacher}, (2) \textit{the student}, (3) \textit{the interaction} and (4) \textit{the environment}. Generally, in this setup teachers transmit language data to their students, according to a certain budget (``a (low) fixed number of bits''). Having this budget forces the teacher to actively choose a learning strategy, as just sending all data that is available to the teacher would not be allowed. Students have the objective to learn the language and they send a signal back that informs their teacher of their performance, e.g., a score on an exam. This interaction takes place in an environment, e.g., a common language. 

In Table~\ref{tab:roadmap} we present the road map that we envision towards interactive language modeling. This road map works as follows. For each of the four aforementioned components we detail steps that need to be taken. We also add a fifth component: the evaluation of the setup. Each component has different aspects (bold-faced in Table~\ref{tab:roadmap}). For example, for the \textit{teacher} we can focus on how it can access the data that it can transmit to the student, which we call ``ways of speaking'' in Table~\ref{tab:roadmap}. Another aspect of the teacher side focuses on what we call the ``degree of awareness'', which entails different ways in which the teacher can remember different aspects of the teaching loop. In a similar fashion we fill in the remaining components in the table. We focus on text as a single modality and acknowledge grounded interactive language modeling as an interesting future research direction.

On our road map there are multiple ways to reach the destination. 
For example, one can focus on taking a few steps for each of the components, or to take many steps for only one or a few of the components. Moreover, although mostly structured in increasing degree of complexity, this does not always hold for all individual steps in the table. For example, zooming in on the ``degrees of awareness'' for the teacher again, one could imagine an example where a teacher does not have an explicit memory buffer of what it sent to the student before, but does have an explicit way of remembering what the student's fine-grained capabilities are, as well as the other way around. 

In the remainder of this work we take the first steps on the road map.
We focus on the teacher side, i.e., learning the correct didactic approach.

\definecolor{Gray0}{gray}{0.6}
\definecolor{Gray1}{gray}{0.8}
\definecolor{Gray2}{gray}{0.95}

\renewcommand{\arraystretch}{1.5}
\begin{table*}
\centering%
\begin{tabularx}{\textwidth}{X  X}
\hline

\rowcolor{Gray2}
\multicolumn{1}{c }{\textbf{Teacher}}       & \multicolumn{1}{c}{\textbf{Student}} \\ 
\hline

\begin{minipage}[t]{\linewidth}
\textbf{Ways of speaking}
  \begin{itemize}[leftmargin=*,nosep]
  \item Select data from bin;
  \item Generate data with own language model.
  \end{itemize}
  \vspace{5pt}
  
\textbf{Degrees of awareness}
  \begin{itemize}[leftmargin=*,nosep]
  \item (No) memory buffer of what has been sent to the student and being able to act on it (see \textit{Interaction} cell);
  \item (No) explicit way of remembering what the student's fine-grained capabilities are and being able to act on it (see \textit{Interaction} cell).
  \end{itemize}
  \vspace{5pt}
\end{minipage}  
& 

\begin{minipage}[t]{\linewidth}
\textbf{Ways of speaking}
  \begin{itemize}[leftmargin=*,nosep]
  \item Generate language data in a standard LM fashion;
  \item Actively experiment with language generation to elicit direct feedback from the teacher (see also \textit{Interaction} cell).
  \end{itemize}
\textbf{Degrees of using the teacher data}
  \begin{itemize}[leftmargin=*,nosep]
  \item Use all data received from the teacher;
  \item Actively select data that is useful;
  \item Actively know when to stop training (for example to avoid overfitting).
  \end{itemize}
\vspace{5pt}
 \vspace{5pt}
\end{minipage}       
\\

\hline

\rowcolor{Gray2}
\multicolumn{1}{c }{\textbf{Interaction}}   & \multicolumn{1}{c}{\textbf{Environment}} \\
\hline 

\begin{minipage}[t]{\linewidth}
\textbf{Teacher side}
  \begin{itemize}[leftmargin=*,nosep]
  \item Send all data at once;
  \item Send data in batches, based on student feedback (see below). Batches can be as small as single utterances, after which the student sends an utterance back, like in real human-to-human interaction (see below);
  \item Send (mid-term) exams.
  \end{itemize}
  \vspace{5pt}
 \textbf{Student side}
  \begin{itemize}[leftmargin=*,nosep]
  \item Send a single average exam score back to the teacher;
  \item Send a fine-grained exam score back, e.g.,
    \begin{itemize}[leftmargin=*,nosep]
      \item score per item on the exam set;
      \item (average) scores of different components (tasks) of the exam(s)
     \end{itemize}
  \item Ask for feedback, for example by actively experiment with language generation for the teacher to judge (`generate own exam').
  \end{itemize}
  \vspace{5pt}
 \end{minipage}                                         
& 

\begin{minipage}[t]{\linewidth}
\textbf{Language}
  \begin{itemize}[leftmargin=*,nosep]
  \item Artificial languages, in increasing level of difficulty in terms of complexity, e.g.,
    \begin{itemize}[leftmargin=*,nosep]
      \item random language;
      \item different types of structures;
      \item different vocabulary sizes;
     \end{itemize}
  \item Subset of human language, e.g., in terms of
    \begin{itemize}[leftmargin=*,nosep]
      \item semantics (e.g., different domains)
      \item syntax (e.g., different grammatical structures)
      \item pragmatics
    \end{itemize}
    \item Unrestricted human language.
  \end{itemize}
  \vspace{5pt}
  
\textbf{Task }
\begin{itemize}[leftmargin=*,nosep]
  \item \textit{Teacher:} Learn to select or generate the optimal data such that the student performs well on the exam set (see cell below);
  \item \textit{Teacher:} Learn to adapt to different types of students, e.g., 
      \begin{itemize}[leftmargin=*,nosep]
      \item architectural differences
      \item different prior knowledge (be aware of catastrophic forgetting in neural networks)
     \end{itemize}
  \item \textit{Student:} Learn to adapt to different types of teachers (didactic strategies).
  \end{itemize}
  \vspace{5pt}
 \end{minipage}    
\\

\hline
\rowcolor{Gray2}

\multicolumn{2}{c}{\textbf{Evaluation / Exam}} \\
\hline
\begin{minipage}[t]{0.97\textwidth}
  \textbf{Teacher}
  \begin{itemize}[leftmargin=*,nosep]
  \item Accuracy in selecting the optimal teaching protocol
  \end{itemize}
  \vspace{5pt}
  \textbf{Student (Exam / Feedback for teacher)}
  \begin{itemize}[leftmargin=*,nosep]
  \item General performance, measured in perplexity;
  \item Performance on specific tasks, such as
    \begin{itemize}[leftmargin=*,nosep]
      \item Subset of the data known to the teacher (e.g., specific domain or (grammatical) structure)
      \item BLIMP~\cite{warstadt-etal-2020-blimp-benchmark};
      \item BIG-Bench~\cite{srivastava2022bigbench} (\url{https://github.com/google/BIG-bench}).
    \end{itemize}
  \item Scores either as an average of more fine-grained (see \textit{Interaction} cell).
  \end{itemize}
  \vspace{5pt}
\end{minipage}   
\\

\hline

\end{tabularx}%
\caption{Road map to interactive language modeling.}
\label{tab:roadmap}
\end{table*}

%% file: sections/04_our_approach.tex

\section{Taking the First Steps on the Road Map}

\begin{figure}
    \centering
    \includegraphics[width=0.48\textwidth]{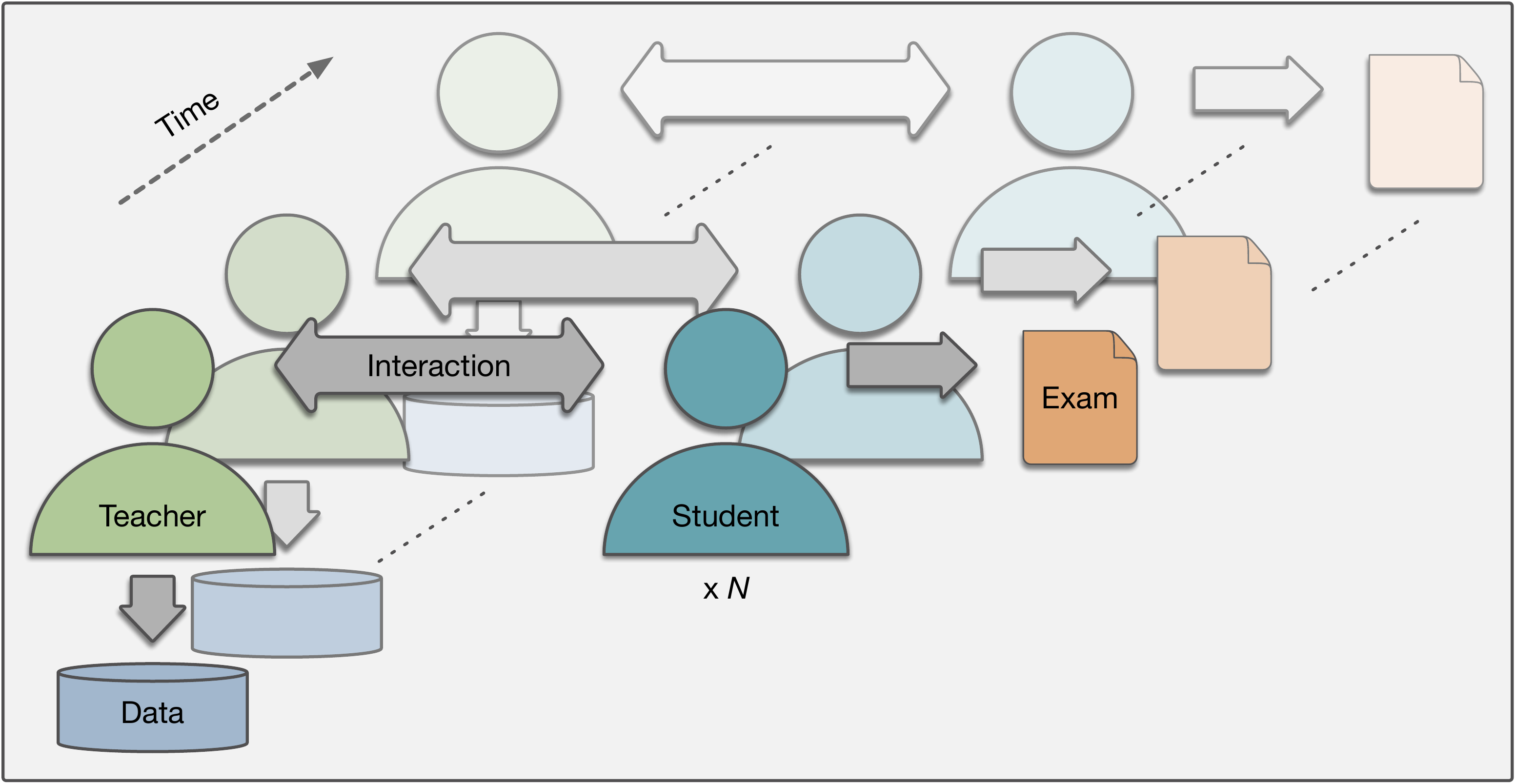}
    \caption{Teacher-student loop as used in this work.}
    \vspace{-1em}
    \label{fig:teacher-student-setup-ours}
\end{figure}

Figure~\ref{fig:teacher-student-setup-ours} shows how we adapt the general setup from Figure~\ref{fig:teacher-student-setup-general} to take the first steps on the road map. Here we describe each modification per component: \textit{the teacher}, \textit{the student}, \textit{the interaction}, \textit{the environment} and \textit{the exam} that the student takes.

\subsection{The Teacher}
In this work we focus on the teacher side. The role of the teacher is to transmit language data that will optimally help the student to learn the language. Figure~\ref{fig:teacher-student-setup-ours} shows that we train the teacher to do this in a number of time steps. At each of these steps a teacher samples data from a larger language data set according to a fixed budget. We discuss the specifics of the sampling function below. To reduce the variance in the teacher's learning process we repeat this process for multiple students, i.e., a teacher selects $N$ ``lessons'' for $N$ students. Due to the stochasticity of the sampling process, each student has the potential to be trained on a slightly different part of the data. Because we use a multiprocessing setup we can train multiple students on a single GPU. Hence, using multiple students does not drastically increase the computational cost. 

\subsubsection{Knowing the Language}
\label{sec:knowing_the_language}
The teacher is modeled as a native speaker of the language that it needs to teach. We represent the teacher's language understanding with a pretrained causal Transformer LM~\cite{vaswani2017attention}. We pretrain this model on a \textit{different} subset of the data than the teacher can select from for the students, and thus we ensure that we measure whether a teacher can teach a language as a whole, and not only a particular subset that it was trained on itself.

\subsubsection{Selecting the Data}
We use REINFORCE~\cite{williams1992simple} with entropy regularization~\cite{mnih2016asynchronous} to learn the teacher's didactic approach.\footnote{We also experimented with gradient-free optimization approaches such as the ones implemented in Nevergrad~\cite{nevergrad}, but found REINFORCE to be more flexible in our case and therefore a better fit for our needs.} We want to optimize the teacher's policy such that it learns to select the optimal data to train the student on, given a predefined budget. The policy is a one-layer feed forward neural network, that outputs a score for each sentence, i.e., the teacher's policy network takes a sentence embedding as input, based on the pretrained Transformer LM that we use to represent the teacher's language understanding. An action is modeled as selecting $k$ sentences from the larger data set, where $k$ is a predefined teacher budget. We use the GumbelTopK trick~\cite{vieira2014gumbel, kool2019stochastic} to sample $k$ sentences without replacement, based on the teacher policy's output scores. We compute the log probabilities (needed to compute the loss) for each sample by adding the log probabilities of each element in the sample. We explain the rationale behind this in Appendix~\ref{sec:appendix_logprobs}.

\subsection{The Student}
As the teacher is the main focus of our work, we choose to keep the student side simple. We represent the student as a causal Transformer LM, that we train on the data that it receives from the teacher.

\subsection{The Interaction}
Following Table~\ref{tab:roadmap}, the teacher sends all selected data to the student at once. The student uses this data to train its LM and takes an exam after a predefined number of updates. The average exam score is sent back to the teacher as feedback. 
We use the student's last model checkpoint to compute the scores (as opposed to the best checkpoint on a validation set), to ensure that the learning signal for the teacher is restricted to the student's performance on the exam set, i.e., we do not expect teachers to reverse the learning process of the students (just like caregivers cannot do this for their children).

\subsection{The Environment}
Following Table~\ref{tab:roadmap}, we design a number of artificial languages to test our approach on (see Section~\ref{sec:experiments} for details). Using artificial languages is a well-tested approach to study the behavior of neural networks~\cite[e.g.,][]{batali1994artificial,wiles1995learning,rodriguez1999recurrent,gers2001lstm,rodriguez2001simple, hupkes2018visualisation,lake2018generalization, saxton2019analysing,hupkes2020compositionality, rodriguez-luna-etal-2020-internal, van-der-wal-etal-2020-grammar,chaabouni2021can,dagan-etal-2021-co}.
Using artificial languages gives us the control we need to design our experiments in such a way that we can correctly interpret the results.

\subsection{The Exam}
The exam is a held-out set over which we compute the student's perplexity. The details of the exam are task dependent and we discuss these next.

%% file: sections/05_experimental_setup.tex

\section{Experiments}
\label{sec:experiments}

We test our proposed setup on a number of settings and tasks, that we describe in this section.

\subsection{Task 1 -- Teaching Different Domains}

For this task we design a language consisting of two strictly separated vocabularies, loosely representing two different domains in natural language. Specifically, $V_1 = \{a, b, c, d, e, f, g, h, i, j\}$, and $V_2 = \{k, l, m, n, o, p, q, r, s, t\}$. We construct sentences by randomly sampling from these sets. Sentences consist either of tokens only from $V_1$ or of tokens only from $V_2$. Sentences have an equal length of $10$ tokens each. Half of the data set that the teacher can choose from consists of $V_1$ sentences, the other half consists of $V_2$ sentences. The teacher's LM is trained on a similarly constructed data set, yet consisting of different sentences. The student’s exam set consists of sentences from only one of the vocabularies, $V_1$ in our case. These are different sentences than in the training set, i.e., the teacher cannot simply sample the exam set to train the student. Hence, the optimal teaching strategy is to present the student with sentences from the exam vocabulary. We confirm this in our baseline experiments that we present in Section~\ref{sec:baseline_experiments}.

\subsection{Task 2 -- Teaching Different Structures}

For this task we do not use different vocabularies, but different sentence structures. All our sentences are constructed with $V_1$ and are between $2$ and $10$ tokens long. We use two different structures: single repetitions and double repetitions. In the case of the single repetitions two identical tokens never occur next to each other, whereas in the case of double repetitions tokens are sampled in pairs:

\begin{enumerate}[leftmargin=*,label=\textit{Structure \arabic* -} , nosep]
\vspace{1pt}
    \item Single repetitions: $(x y) ^ n$
    \item Double repetitions: $(x x)$ or $(x x y y) ^ n$
\end{enumerate}

\noindent The data set that the teacher can sample from consists for $20\%$ of sentences with Structure $1$ and for $80\%$ of sentences of Structure $2$. The exam set consists of sentences with Structure $1$. We opt for this way of splitting the data, as we found that a student performs quite well when trained on data consisting half of Structure $1$ and half of Structure $2$. Having an unequal split thus allows us to make sure that we can appropriately distinguish a learned didactic approach from a random one. For this task the optimal teaching strategy is to select sentences with the exam structure, as we confirm with our baseline experiments that we present in Section~\ref{sec:baseline_experiments}.

\subsection{Training Details}
\label{sec:training_details}
The teacher LM is trained on $100$ unique sentences till convergence. The dataset the teacher can sample from for the student consists of $100$ different unique sentences.
The exam consists of $10$ unique sentences and we set the teacher budget to $10$ as well.
We run our experiments with five different random seeds and report the averages and standard deviations. We use the negative perplexity of the student on the exam as reward for the teacher. We experiment with two different sentence embeddings for the teacher: average word embeddings and the average of the last hidden layer. We train students for a predefined number of steps that we determine by inspecting the loss and perplexity curves of training an LM once before the actual experiments. We base the threshold on when a student LM starts to overfit, so that a teacher can get clear feedback signals. We set this value to $400$ for Task $1$ and $300$ for Task $2$. Automatically determining when the students stops training is an important avenue for future work (Table~\ref{tab:roadmap}). We use Fairseq's~\cite{ott-etal-2019-fairseq} \texttt{transformer\_lm}\footnote{\url{https://fairseq.readthedocs.io/en/latest/command_line_tools.html}} for the implementation of the Transformer LMs. We use up to four GPUs with 32 GB RAM per experiment. The exact number depends on the number of students per teacher, as we can fit up to $6$ students on a single GPU due to our multiprocessing implementation.

\subsection{Baseline experiments}
\label{sec:baseline_experiments}
We run three baseline experiments with three different didactic strategies:
an \textit{oracle}, \textit{random}, and \textit{worst case} strategy. We run the baselines for five different random seeds. 
In each experiment, we randomly select data according to the teacher budget. We do this five times and each time train a student LM with the selected data. The difference between baselines is the type of data that can be selected. For the oracle baseline we only select sentences that consist of the exam vocabulary (Task $1$) or structure (Task $2$). For the random baseline we randomly select sentences. For the worst case baseline all sentences that we select are from a different vocabulary or structure than the exam sentences.

%% file: sections/06_results.tex

\section{Results}

\subsection{Task 1 -- Different Domains}

\subsubsection{Baseline Results}
In Table~\ref{tab:baseline_results_task_1} we present the results for the baseline experiments for Task 1. 
We report the averages and standard deviations of the perplexity on the exam set and the fraction of training sentences that consisted of the exam vocabulary. For space reasons, we report the results for two seeds per baseline: the seed with the best average perplexity and the worst. The results for all fives seeds are given in Appendix~\ref{sec:appendix_baseline_results_task1}. There we also present scores for the $n$-gram overlap between the selected training set and the exam set. The results are as expected. The oracle baseline gives the best results, followed by the random and worst case baseline respectively.

\begin{table}
\centering
\renewcommand{\arraystretch}{1}

\begin{tabular}{crrr}
\toprule
\textbf{Type} & \textbf{Seed} & \multicolumn{1}{c}{\textbf{Avg}}   & \multicolumn{1}{c}{\textbf{Avg train}}   \\

& & \multicolumn{1}{c}{\textbf{Perplexity}} & \multicolumn{1}{c}{\textbf{from test}} \\ 
\midrule

\textit{Rand.} & B & $160.9 \pm 217.7 $ & $0.54 \pm 0.16 $ \\
                & W & $742.5 \pm 159.8 $ & $0.50 \pm 0.17 $ \\

\midrule

\textit{Orac.} & B & $14.99 \pm 5.364 $ & $1.00 \pm 0.00$ \\
                & W & $68.95 \pm 87.49 $ & $1.00 \pm 0.00$ \\

\midrule

\textit{Worst} & B & $4.78e4 \pm 2.67e4$ & $0.00 \pm 0.00$ \\
\textit{case}  & W & $8.46e4 \pm 4.69e4$ & $0.00 \pm 0.00$ \\

\bottomrule
\end{tabular}
\caption{Baseline results Task $1$. Averages and standard deviations reported based on five runs per seed. \textit{Rand} is Random, \textit{Orac} is Oracle, \textit{B} is Best and \textit{W} is Worst.}
\vspace{-1em}
\label{tab:baseline_results_task_1}
\end{table}

\subsection{Results of Training the Teacher}

\begin{figure*}[htp]
     \centering
     \begin{subfigure}[b]{0.4\textwidth}
         \centering
         \includegraphics[width=\textwidth]{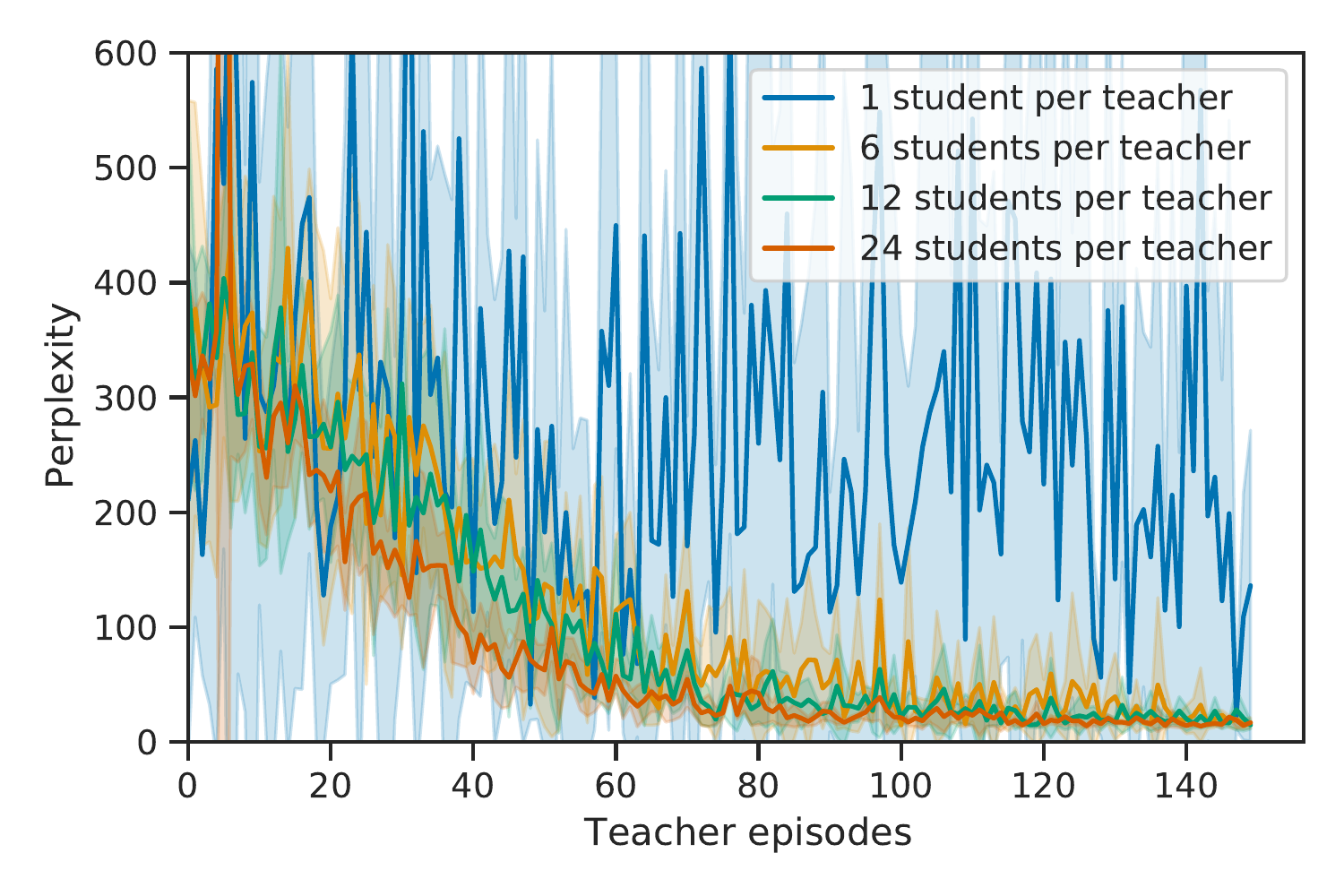}
         \vspace{-2em}
         \caption{Perplexity of the student on the exam data over different episodes. Average word embedding as input to the teacher's policy.}
         \label{fig:exp2_ppl_avg_word}
     \end{subfigure}
     \hfill
     \begin{subfigure}[b]{0.4\textwidth}
         \centering
         \includegraphics[width=\textwidth]{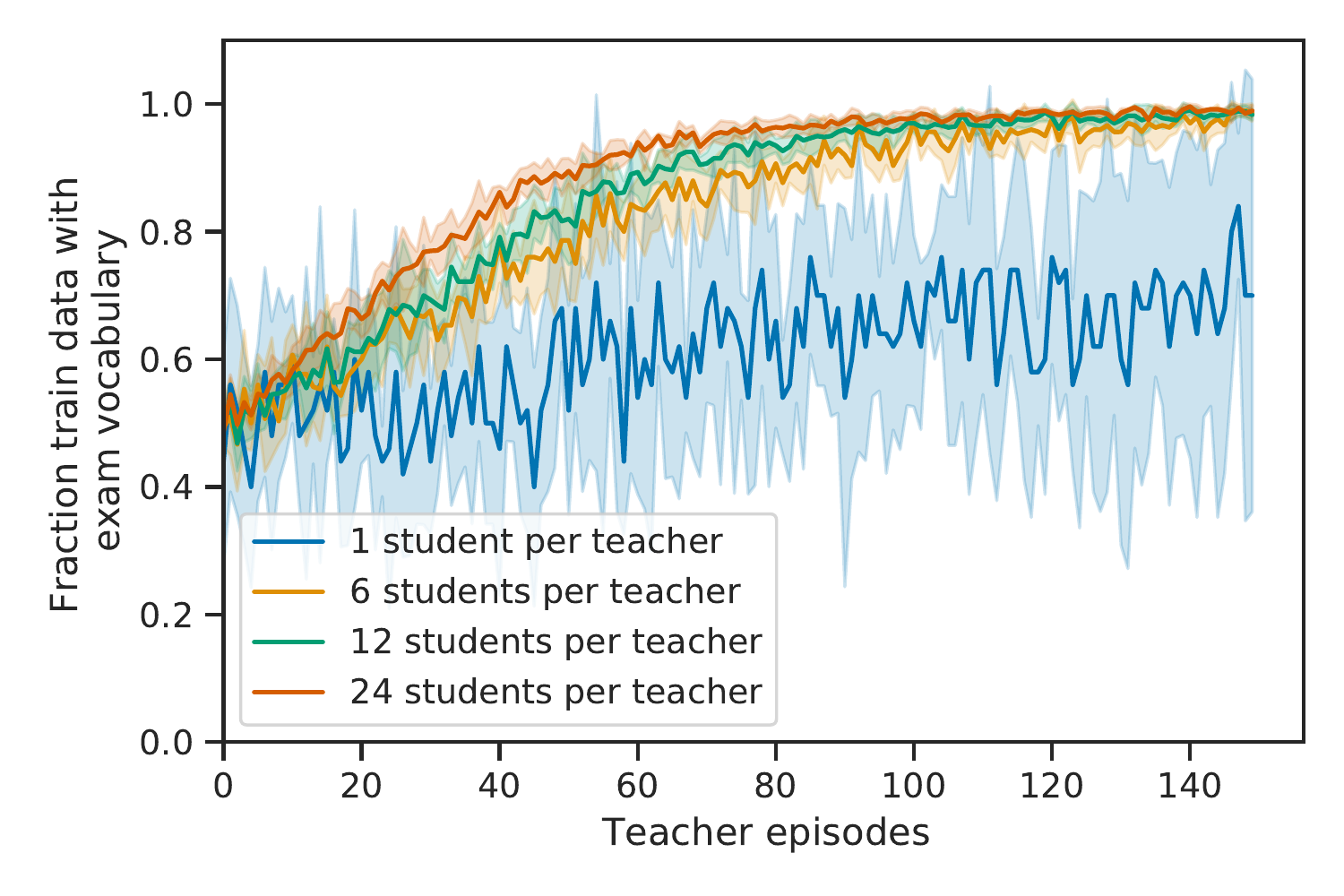}
         \vspace{-2em}
         \caption{Fraction training data with the exam vocabulary over different episodes. Average word embedding as input to the teacher's policy.}
         \label{fig:exp2_train_from_test_avg_word}
     \end{subfigure}
          \begin{subfigure}[b]{0.4\textwidth}
         \centering
         \includegraphics[width=\textwidth]{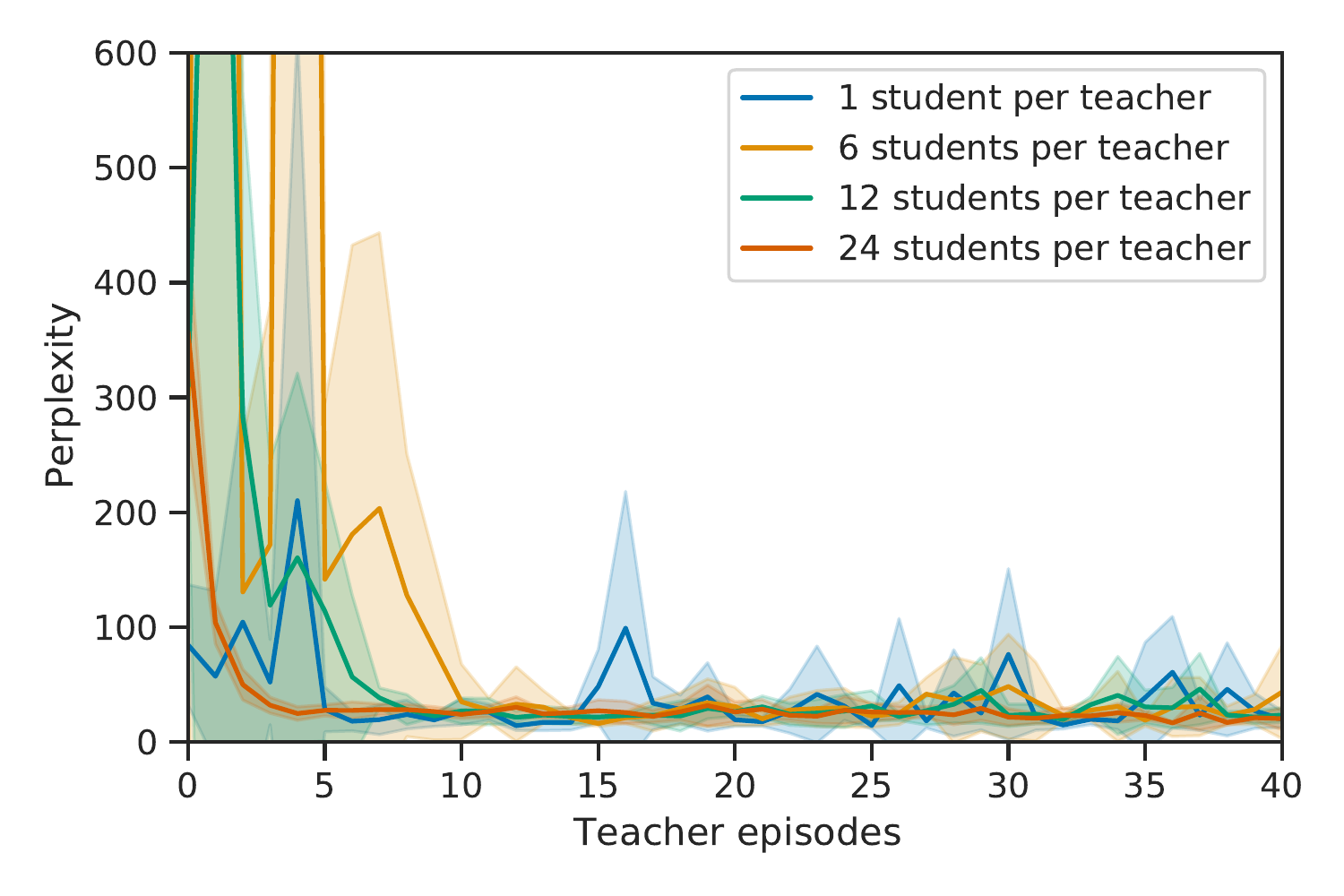}
         \vspace{-2em}
         \caption{Perplexity of the student on the exam data over different episodes. Average last hidden layer as input to the teacher's policy.}
         \label{fig:exp2_ppl_avg_hidden}
     \end{subfigure}
     \hfill
     \begin{subfigure}[b]{0.4\textwidth}
         \centering
         \includegraphics[width=\textwidth]{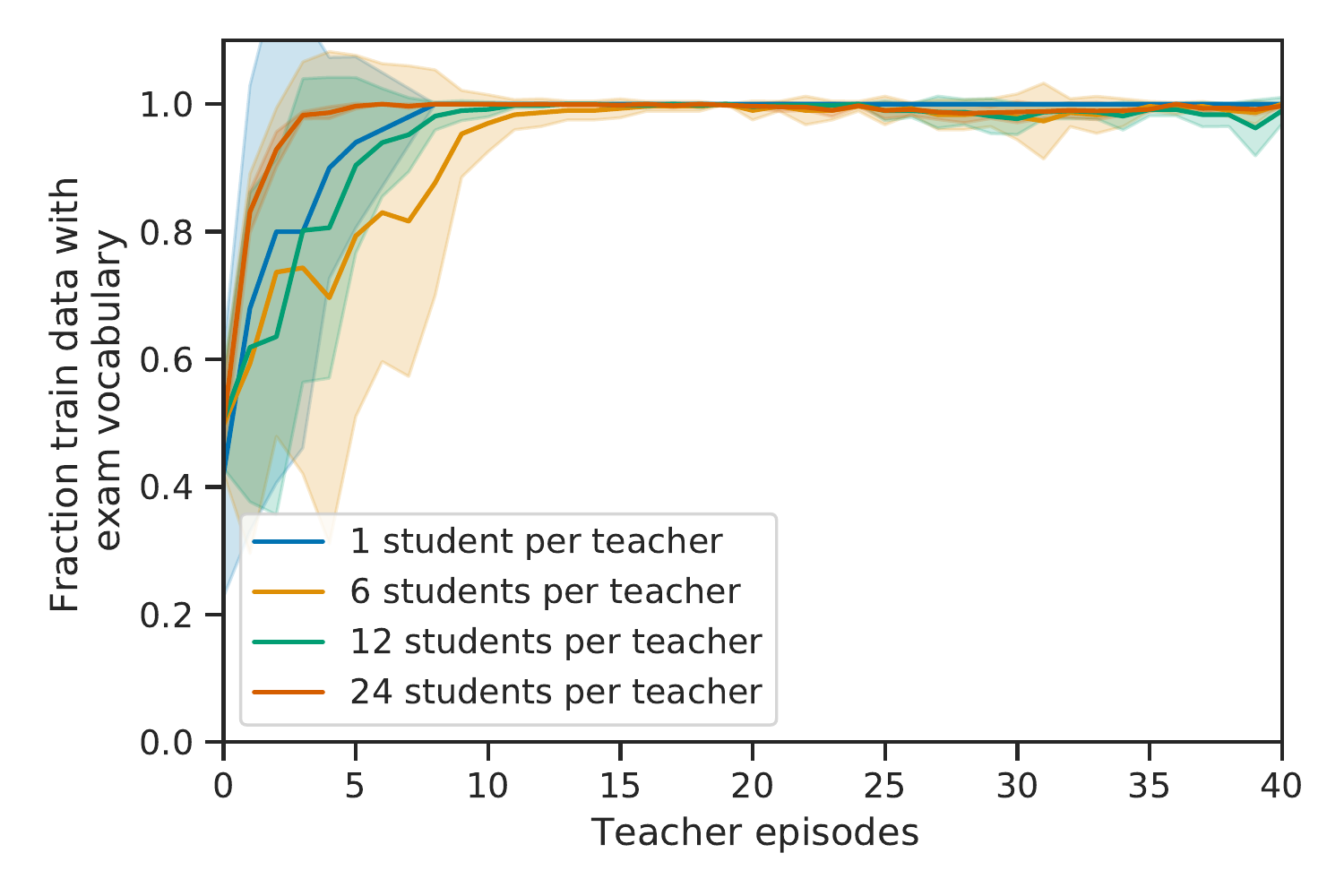}
         \vspace{-2em}
         \caption{Fraction training data with the exam vocabulary over different episodes. Average last hidden layer as input to the teacher's policy.}
         \label{fig:exp2_train_from_test_avg_hidden}
     \end{subfigure}
     \hfill
        \caption{Results Task 1 -- Different domains. Plots for different numbers of students per teacher. Results per setting reported as average and standard deviation over five random seeds. x-axis of lower plots bound to $40$ as the teacher had already converged by then.}
        \vspace{-1em}
 
        \label{fig:results_task1}
\end{figure*}

\begin{figure*}[htp]
     \centering
     \begin{subfigure}[b]{0.24\textwidth}
         \centering
         \includegraphics[width=\textwidth]{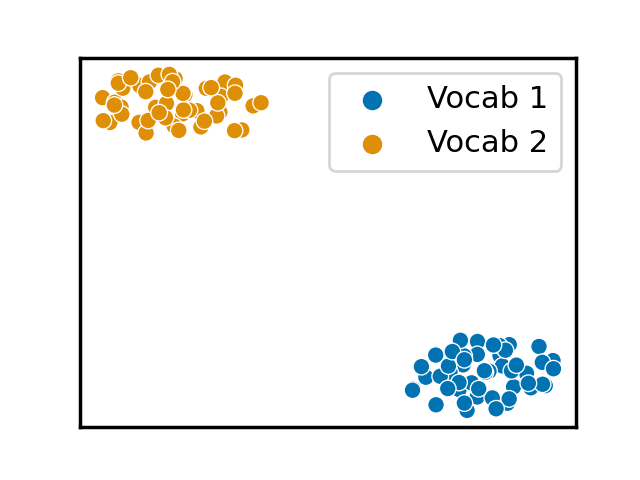}
         \caption{Task 1 - Different vocabularies. Sentence embedding is average word embeddings.}
         \label{fig:tsne_task1_avg_word}
     \end{subfigure}
     \hfill
     \begin{subfigure}[b]{0.24\textwidth}
         \centering
         \includegraphics[width=\textwidth]{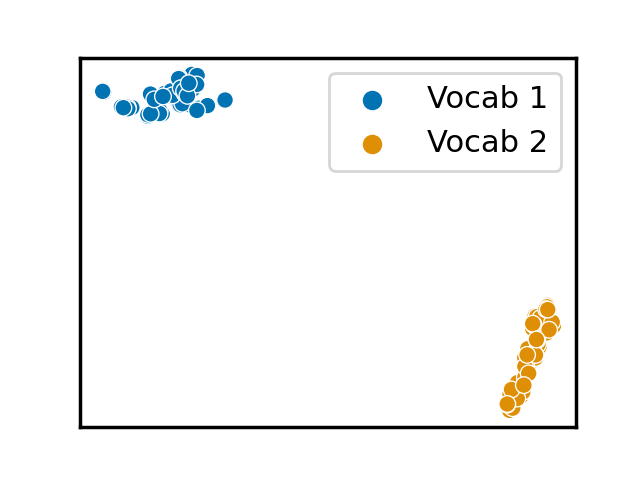}
         \caption{Task1 - Different vocabularies. Sentence embedding is average last hidden layer.}
         \label{fig:tsne_task1_avg_hidden_layer}
     \end{subfigure}
    \begin{subfigure}[b]{0.24\textwidth}
         \centering
         \includegraphics[width=\textwidth]{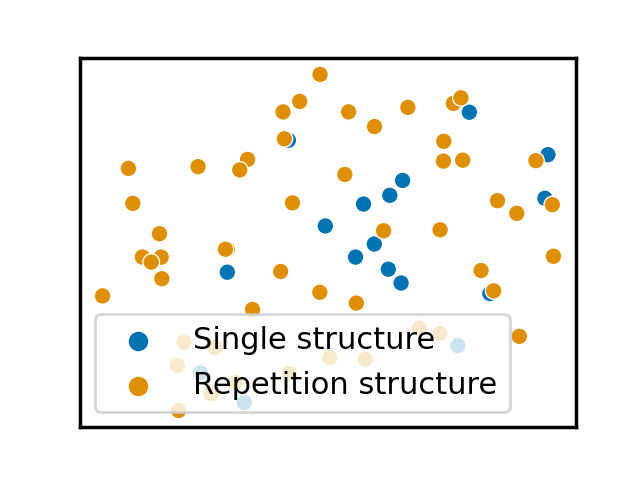}
         \caption{Task 2 - Different structures. Sentence embedding is average word embeddings.}
         \label{fig:tsne_task2_avg_word}
     \end{subfigure}
     \hfill
     \begin{subfigure}[b]{0.24\textwidth}
         \centering
         \includegraphics[width=\textwidth]{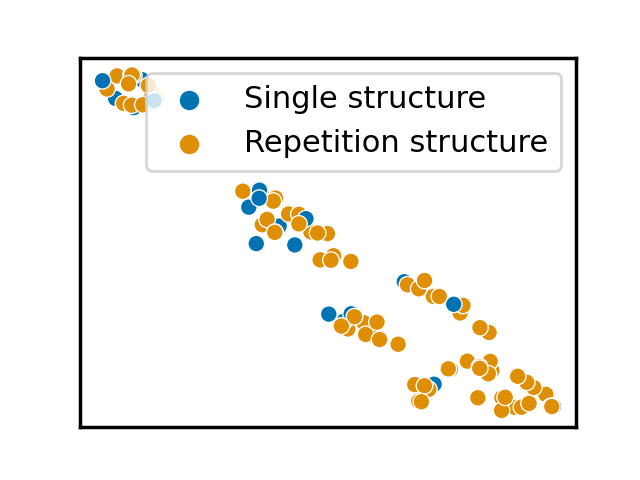}
         \caption{Task 2 - Different structures. Sentence embedding is average last hidden layer.}
         \label{fig:tsne_task2_avg_hidden_layer}
     \end{subfigure}
     \hfill
        \caption{T-SNE plots for different sentence representations for different tasks.}
        \vspace{-2em}
        \label{fig:tsne_sent_embeddings}
\end{figure*}

In Figure~\ref{fig:results_task1} we present the results for Task $1$ for different numbers of students per teacher.\footnote{We present plots for the $n$-gram overlap in Appendix~\ref{sec:appendix_results_task1}.} The teacher's didactic strategy correctly converges to the oracle baseline.
There is a clear difference between different sentence embeddings (Section~\ref{sec:knowing_the_language}).  
Both embedding types are converging, but the average hidden layer embeddings are clearly superior. We investigate this further by plotting the t-SNE embeddings~\cite{van2008visualizing} of the different sentence embeddings in Figure~\ref{fig:tsne_sent_embeddings}. To prepare for Task $2$, we also plot the embeddings of Task $2$. The hidden layer sentence embeddings result in the clearest separation between sentences from different vocabularies or structures. Especially for Task $2$, where we use the same vocabulary, this is unsurprising. From now on we opt for these sentence embeddings. Based on the results for Task 1 we opt for $12$ students per teacher as a good trade-off between computational cost and convergence stability for Task $2$.

\begin{figure*}[htp]
     \centering
     \begin{subfigure}[b]{0.4\textwidth}
         \centering
         \includegraphics[width=\textwidth]{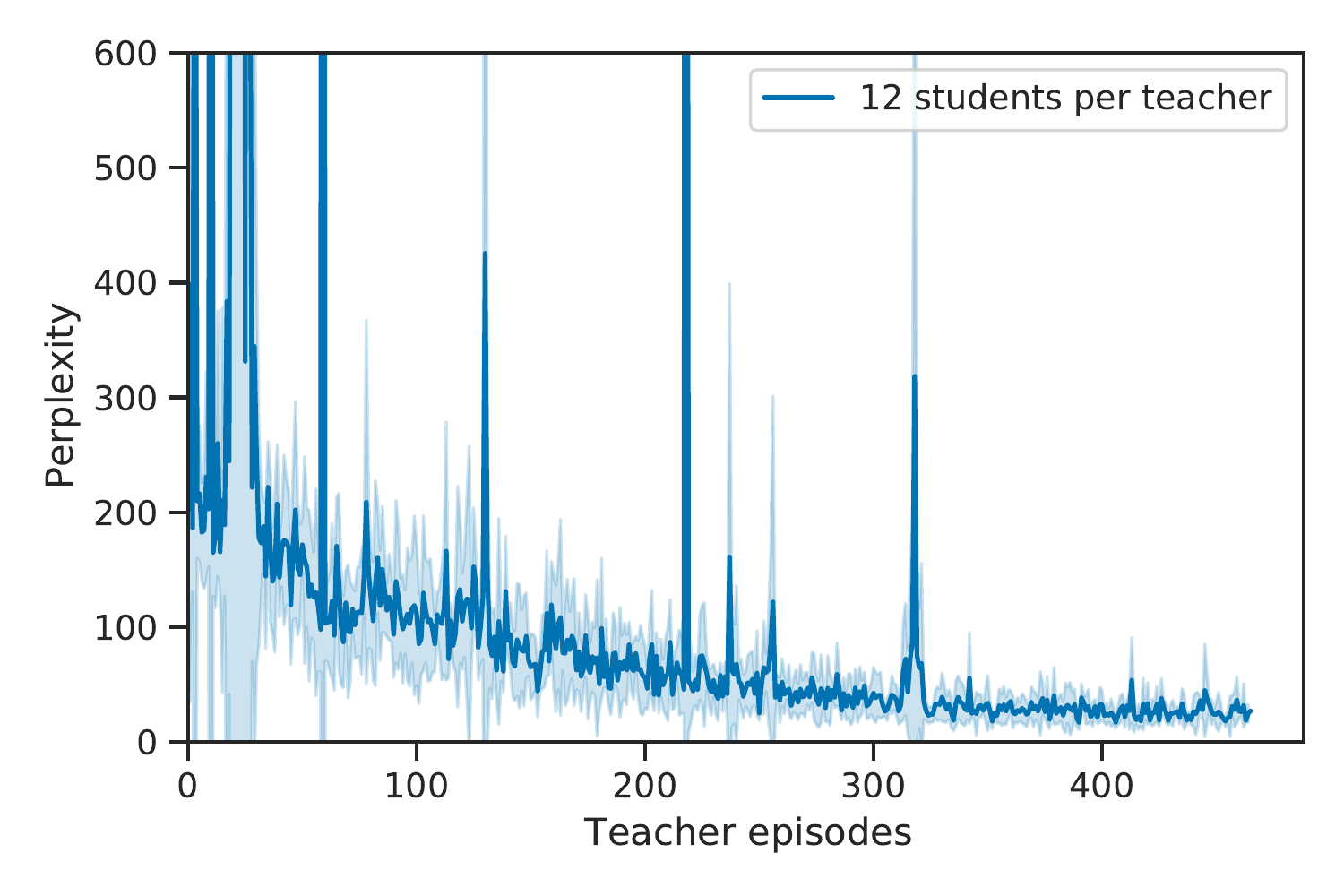}
         \vspace{-2em}
         \caption{Perplexity of the student on the exam data over different episodes.}
         \label{fig:exp2_ppl}
     \end{subfigure}
     \hfill
     \begin{subfigure}[b]{0.4\textwidth}
         \centering
         \includegraphics[width=\textwidth]{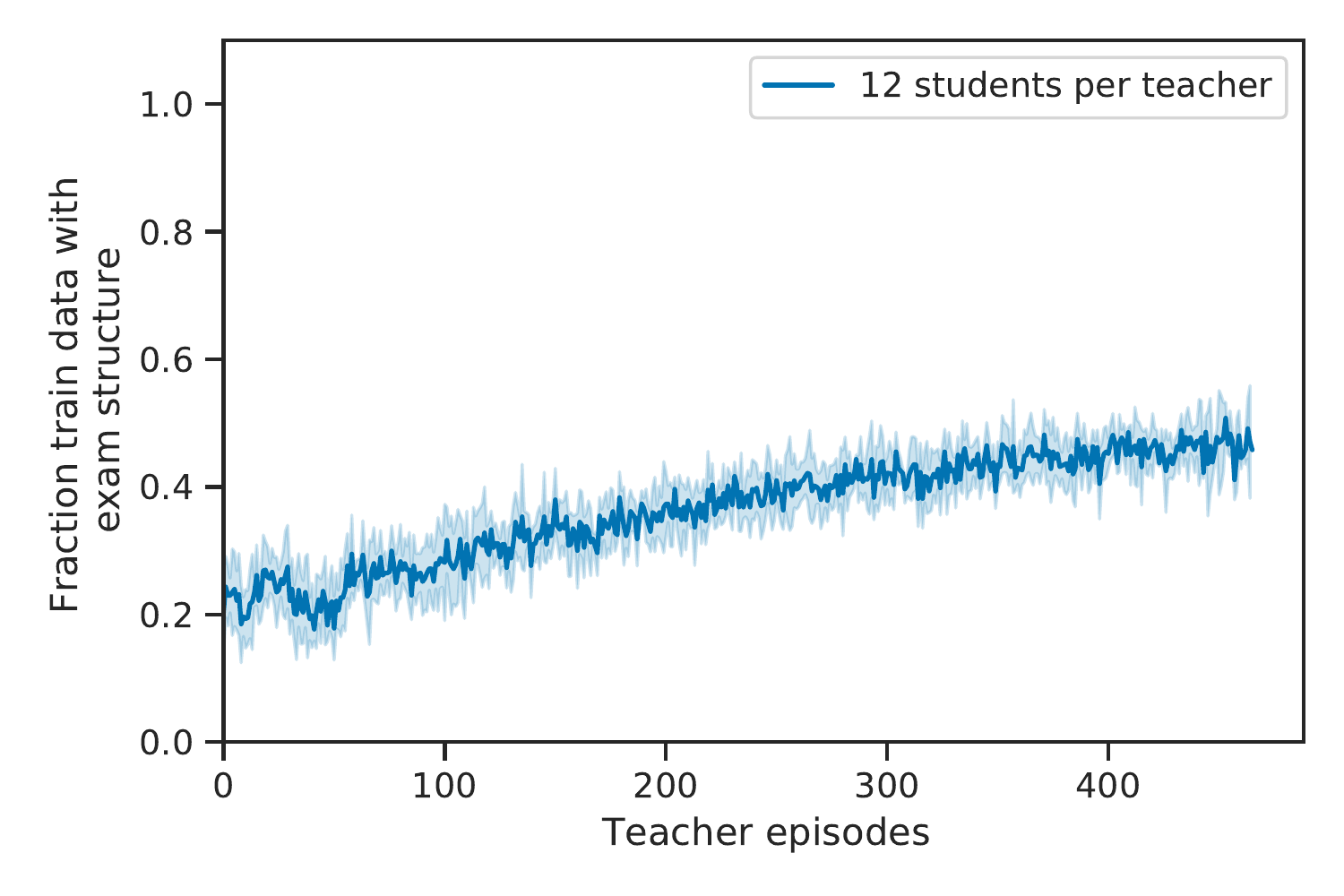}
         \vspace{-2em}
         \caption{Fraction training data with the exam structure over different episodes.}
         \label{fig:exp2_train_from_test}
     \end{subfigure}
     \hfill
        \caption{Results Task 2 -- Plots for 12 students per teacher. Results per setting reported as average and standard deviation over five random seeds.}
        \label{fig:results_task2}
        \vspace{-2em}
\end{figure*}

\subsection{Task 2 -- Different Structures}
\label{sec:experiments_task2}

\subsubsection{Baseline Results}

We present the results for the baseline results for Task $2$ in Table~\ref{tab:baseline_results_task_2}. Again we report the results for the best and the worst seed. Full results are available in Appendix~\ref{sec:appendix_baseline_results_task2}. Similarly to the results for Task $1$, we confirm that the oracle baseline performs strongest, followed by the random and worst case baseline respectively.

\begin{table}
\centering
\renewcommand{\arraystretch}{1}

\begin{tabular}{crrr}
\toprule
\textbf{Type} & \textbf{Seed} & \multicolumn{1}{c}{\textbf{Avg}}   & \multicolumn{1}{c}{\textbf{Avg train}} \\
& & \multicolumn{1}{c}{\textbf{Perplexity}} & \multicolumn{1}{c}{\textbf{from test}} \\ 
\midrule

\textit{Rand.} & B & $119.0 \pm 56.48 $ & $0.18 \pm 0.04$  \\
                & W & $342.1 \pm 241.4 $ & $0.12 \pm 0.08$  \\
\midrule

\textit{Orac.} & B & $6.821 \pm 0.619 $ & $1.00 \pm 0.00$  \\
                & W & $9.431 \pm 3.057 $ & $1.00 \pm 0.00$ \\
                
\midrule

\textit{Worst} & B & $299.6 \pm 124.2$ & $0.00 \pm 0.00$ \\
\textit{Case}  & W & $595.3 \pm 297.9$ & $0.00 \pm 0.00$ \\
                    
\bottomrule
\end{tabular}
\caption{Baseline results Task $2$. Averages and standard deviations reported based on five runs per seed. \textit{Rand} is Random, \textit{Orac} is Oracle, \textit{B} is Best and \textit{W} is Worst.}
\label{tab:baseline_results_task_2}
\vspace{-1em}
\end{table}

\subsubsection{Results of Training the Teacher}

In Figure~\ref{fig:results_task2} we present the results for Task $2$.\footnote{We present plots for the $n$-gram overlap in Appendix~\ref{sec:appendix_results_task2}.} Again we see that the teacher learns to gradually converge to the oracle teaching strategy, although convergence is less fast than for Task $1$; we do not achieve full convergence in the number of training episodes that we run these experiments for. We postulate that this can be explained by the differences we found in Figure~\ref{fig:tsne_sent_embeddings}. The differences in sentence embeddings between the two different structures are clearly less apparent than between the sentences from two vocabularies. This indicates the importance of good sentence embeddings in future work. Moreover, as stated in Section~\ref{sec:experiments_task2}, we found that transmitting roughly $50\%$ of Structure $1$ and $50\%$ of Structure $2$ also already leads to good performance.
Therefore, the teacher likely needs to learn from a less distinct learning signal than in Task $1$. 

%% file: sections/07_implications.tex

\section{Implications and Outlook}
\label{sec:implications}

We successfully took the first steps on our proposed road map.
Here we want to share our learnings and the limitations of the current setup to help future research to take the next steps on the road map.
\\

\noindent \textbf{The importance of designing experiments with interpretable outcomes.} We designed our experiments such that we knew the teacher's oracle strategy, which allowed us to properly test our setup. However, in designing our experiments we found that finding such settings is non-trivial. For example, in a task that contains a language with multiple structures, a student might unexpectedly learn information from structure $1$ that also proves useful for structure $2$. This might be acceptable if one's only objective is to obtain a good performance. However, in our case it is critical to be able to know that a teacher is ``right for the right reasons'', which motivated our choices for the tasks and languages.

\noindent \textbf{The teacher's budget.} Following our objective, we designed our experiments in such a way that the teacher was given a budget that limits the amount of data it can send to the student. As mentioned in Section~\ref{sec:training_details}, we confirmed that the student's learning converges with this budget. In follow up work we plan to investigate the importance of different budgets in more detail. One interesting direction is to give the teacher a flexible budget, i.e., such that a teacher could decide to stop training if it deems it no longer necessary for the student.

\noindent \textbf{Computational complexity.} Apart from the multiprocessing setup that allows us to train multiple students on a single GPU, we did not yet focus on the computational complexity of our approach. In the current setup many student language models need to be trained for a single teacher. In our case we deem this justifiable as we are just at the start of the road map. Moreover, once a teacher model is trained, it can be used for many different purposes. However, in future work we hope to focus on decreasing the computational complexity of our approach. One promising avenue to do this is by optimizing the learning process of the student.

%% file: sections/08_conclusion.tex

\section{Conclusion}
In this paper we pioneered the space of interactive language modeling, motivated by the observation that current state-of-the-art LMs are trained in a very unnatural way, from the perspective of human language acquisition. Moreover, an interactive approach has the potential to make LMs more efficient and adaptable. Specifically, we proposed a teacher-student loop, in which the teacher is inspired by the caregiver and the student resembles the child in the human language acquisition. We presented a road map that details the steps towards interactive language modeling for each of the components of the teacher-student loop. We led by example and took the first steps on this road map, leading to a tangible proof of concept of our proposal. As such, we structured the space of interactive language modeling and aim to inspire a larger research agenda on interactive language modeling.

%% file: sections/09_ethical_impact.tex

\section{Ethical Impact Statement}

At this point we use artificial language data only, for which we do not see any direct negative implications. As we move towards using real data sets, it is necessary to be aware of potential biases with these data sets. One needs to ensure that the data is not biased towards any (protected) group to avoid any harm. Currently, much of the NLP research focuses on English as its language of interest. Our approach is not bound to any language in particular and can even be used to improve language learning in a low resource setting. Once the models achieve human like performance and are used for downstream tasks and applications it is necessary to explicitly state that language is produced by an artificial language model. However, as with all language models, misuse can still happen and it is our responsibility as a research community, amongst others, to spend effort on making users aware of these possibilities.

%% file: sections/10_appendix.tex

\section{Computing the Probability of a Top-K Sample}
\label{sec:appendix_logprobs}

Our objective is to find the (log) probability of sampling the subset $(i_1, ... i_K)$ from $\{1, ..., N\}$ \textit{without} replacement from the categorical probability $(p_1, ..., p_N)$.
\\

\noindent Let us first consider sampling $K$ elements from the $\{1, ..., N\}$ \textit{with} replacement. In that case

\begin{equation}
    p(i_1, ..., i_K) = \prod_{k=1}^K p_{i_k}.
\end{equation}

\noindent If we allow for all possible permutations of observing $(i_1, ..., i_K)$ we get

\begin{equation}
\label{eq:sampling_with_replacement}
    p(i_1, ..., i_K) = C \prod_{k=1}^K p_{i_k},
\end{equation}

where $C = K!$.
\\

\noindent To go from sampling \textit{with} replacement, to sampling \textit{without} replacement, we consider event $A = \textrm{``all sampled elements } (i_1, ..., i_K) \textrm{ are unique''}$. Then

\begin{equation}
\begin{split}
    p_{\textrm{w/o replacement}}(i_1, ..., i_K) = \\
    p_{\textrm{w/ replacement}}(i_1, ..., i_K | A). 
\end{split}
\end{equation}

\noindent Applying Bayes Rule gives us:

\begin{equation}
\begin{split}
    p_{\textrm{w/o replacement}}(i_1, ..., i_K) = \\
    \frac{p_{\textrm{w/ replacement}}(A | i_1, ..., i_K) p_{\textrm{w/ replacement}}(i_1, ..., i_K)}{p_{\textrm{w/ replacement}}(A)}.
\end{split}
\end{equation}

\noindent As in our case all samples in $(i_1, ..., i_K)$ are unique we know that

\begin{equation}
    p_{\textrm{w/ replacement}}(A | i_1, ..., i_K) = 1.
\end{equation}

\noindent Combining this with Equation~\ref{eq:sampling_with_replacement} gives us

\begin{equation}
    p_{\textrm{w/o replacement}}(i_1, ..., i_K) = \frac{C \prod_{k=1}^K p_{i_k}}{p(A)},
\end{equation}

\noindent and thus

\begin{equation}
    p_{\textrm{w/o replacement}}(i_1, ..., i_K) \propto \prod_{k=1}^K p_{i_k},
\end{equation}

\noindent and 

\begin{equation}
    \label{eq:log_probs}
    \log{p_{\textrm{w/o replacement}}(i_1, ..., i_K)} \propto \sum_{k=1}^K \log p_{i_k}.
\end{equation}

\noindent From an implementation perspective this this boils down to the following steps:

\begin{enumerate}[leftmargin=*,nosep]
    \item We compute the scores per sentence.
    \item We sample $K$ sentences without replacement, using the GumbelTopK trick.
    \item We compute the log probabilities for each score: $\log{\textrm{softmax}(scores)}$.
    \item We compute the log probability of our sample by adding the log probabilities of the elements in our sample, according to Equation~\ref{eq:log_probs}.
\end{enumerate}

\subsection{Comparison to Prior Work}
Our problem of sampling $K$ sentences as a single action is similar to the problem formulation of using Reinforcement Learning for extractive summarization to optimize for Rouge~\cite{lin-2004-rouge} directly. In this setting $K$ sentences need to be selected from a document. This results in a very large search space. \citet{narayan-etal-2018-ranking} limit the search space by first selecting $n$ sentences that have a high Rouge score. Then all possible summaries are made with these $n$ sentences. These summaries are ranked according to their Rouge scores and the top $K$ sentences are taken as action. This approach has the disadvantage that it limits the search space heuristically, which does not guarantee that the best summary is found. \citet{dong-etal-2018-banditsum} frame the problem as a contextual bandit problem, which allows them to sample from the true action space. We choose our approach as it is intuitive, simple and effective.

\onecolumn

\section{Additional Results Baseline Experiments Task 1}
\label{sec:appendix_baseline_results_task1}

In Table~\ref{tab:app_baseline_results_task_1} we present the results for our baseline runs on all five seeds.

\begin{table*}[h]
\centering
\renewcommand{\arraystretch}{1}
\begin{tabular}{lrrrrrr}
\toprule
\textbf{Baseline} & \textbf{Seed} & \multicolumn{1}{c}{\textbf{Avg}}   & \multicolumn{1}{c}{\textbf{Avg train}} & \multicolumn{1}{c}{\textbf{Avg unigram}} & \multicolumn{1}{c}{\textbf{Avg bigram}} & \multicolumn{1}{c}{\textbf{Avg trigram}}     \\

& & \multicolumn{1}{c}{\textbf{Perplexity}} & \multicolumn{1}{c}{\textbf{from test}} & \multicolumn{1}{c}{\textbf{overlap}} & \multicolumn{1}{c}{\textbf{overlap}} & \multicolumn{1}{c}{\textbf{overlap}} \\ 
\midrule
\textit{Random} & $6639$ & $193.9 \pm 100.3 $ & $0.46 \pm 0.14 $ & $0.46 \pm 0.14 $ & $0.278 \pm 0.07 $ & $0.023 \pm 0.009 $ \\
                & $7519$ & $683.1 \pm 634.3 $ & $0.52 \pm 0.15 $ & $0.52 \pm 0.15 $ & $0.291 \pm 0.10 $ & $0.030 \pm 0.010 $ \\
                & $1007$ & $742.5 \pm 159.8 $ & $0.50 \pm 0.17 $ & $0.50 \pm 0.17 $ & $0.298 \pm 0.10 $ & $0.035 \pm 0.014 $ \\
                & $4520$ & $160.9 \pm 217.7 $ & $0.54 \pm 0.16 $ & $0.54 \pm 0.16 $ & $0.327 \pm 0.09 $ & $0.035 \pm 0.025 $ \\
                & $4527$ & $307.1 \pm 295.1 $ & $0.58 \pm 0.17 $ & $0.58 \pm 0.17 $ & $0.349 \pm 0.10 $ & $0.035 \pm 0.014 $ \\
\midrule
\textit{Oracle} & $6639$ & $14.99 \pm 5.364 $ & $1.00 \pm 0.00$ & $1.00 \pm 0.00$ & $0.551 \pm 0.06$ & $0.072 \pm 0.029 $ \\
                & $7519$ & $44.37 \pm 58.94 $ & $1.00 \pm 0.00$ & $1.00 \pm 0.00$ & $0.611 \pm 0.02$ & $0.085 \pm 0.017 $ \\
                & $1007$ & $68.95 \pm 87.49 $ & $1.00 \pm 0.00$ & $1.00 \pm 0.00$ & $0.598 \pm 0.02$ & $0.077 \pm 0.025 $ \\
                & $4520$ & $15.65 \pm 4.616 $ & $1.00 \pm 0.00$ & $1.00 \pm 0.00$ & $0.578 \pm 0.02$ & $0.087 \pm 0.028 $ \\
                & $4527$ & $23.66 \pm 21.44 $ & $1.00 \pm 0.00$ & $1.00 \pm 0.00$ & $0.624 \pm 0.02$ & $0.095 \pm 0.019 $ \\
\midrule
\textit{Worst case} & $6639$ & $8.46e4 \pm 4.69e4$ & $0.00 \pm 0.00$ & $0.00 \pm 0.00$ & $0.00 \pm 0.00$ & $0.00 \pm 0.00$ \\
                    & $7519$ & $7.03e4 \pm 3.73e4$ & $0.00 \pm 0.00$ & $0.00 \pm 0.00$ & $0.00 \pm 0.00$ & $0.00 \pm 0.00$ \\
                    & $1007$ & $8.17e4 \pm 4.26e4$ & $0.00 \pm 0.00$ & $0.00 \pm 0.00$ & $0.00 \pm 0.00$ & $0.00 \pm 0.00$ \\
                    & $4520$ & $4.78e4 \pm 2.67e4$ & $0.00 \pm 0.00$ & $0.00 \pm 0.00$ & $0.00 \pm 0.00$ & $0.00 \pm 0.00$ \\
                    & $4527$ & $6.69e4 \pm 1.98e4$ & $0.00 \pm 0.00$ & $0.00 \pm 0.00$ & $0.00 \pm 0.00$ & $0.00 \pm 0.00$ \\

\bottomrule
\end{tabular}
\caption{Baseline results for Task $1$. Different domains. Averages and standard deviations reported based on five runs per seed.}
\label{tab:app_baseline_results_task_1}
\end{table*}

\section{Additional Results Baseline Experiments Task 2}
\label{sec:appendix_baseline_results_task2}

In Table~\ref{tab:app_baseline_results_task_2} we present the results for our baseline runs on all five seeds. 

\begin{table*}
\centering
\renewcommand{\arraystretch}{1}
\begin{tabular}{lrrrrrr}
\toprule
\textbf{Baseline} & \textbf{Seed} & \multicolumn{1}{c}{\textbf{Avg}}   & \multicolumn{1}{c}{\textbf{Avg train}} & \multicolumn{1}{c}{\textbf{Avg unigram}} & \multicolumn{1}{c}{\textbf{Avg bigram}} & \multicolumn{1}{c}{\textbf{Avg trigram}}     \\

& & \multicolumn{1}{c}{\textbf{Perplexity}} & \multicolumn{1}{c}{\textbf{from test}} & \multicolumn{1}{c}{\textbf{overlap}} & \multicolumn{1}{c}{\textbf{overlap}} & \multicolumn{1}{c}{\textbf{overlap}} \\ 
\midrule
\textit{Random} & $6639$ & $119.0 \pm 56.48 $ & $0.18 \pm 0.04$ & $1.00 \pm 0.00$ & $0.401 \pm 0.033 $ & $0.030 \pm 0.020 $ \\
                & $7519$ & $162.8 \pm 201.9 $ & $0.24 \pm 0.05$ & $1.00 \pm 0.00$ & $0.408 \pm 0.044 $ & $0.035 \pm 0.038 $ \\
                & $1007$ & $234.1 \pm 192.0 $ & $0.24 \pm 0.12$ & $1.00 \pm 0.00$ & $0.414 \pm 0.034 $ & $0.034 \pm 0.020 $ \\
                & $4520$ & $161.7 \pm 190.6 $ & $0.22 \pm 0.04$ & $1.00 \pm 0.00$ & $0.410 \pm 0.023 $ & $0.038 \pm 0.033 $ \\
                & $4527$ & $342.1 \pm 241.4 $ & $0.12 \pm 0.08$ & $1.00 \pm 0.00$ & $0.348 \pm 0.024 $ & $0.013 \pm 0.017 $ \\
\midrule
\textit{Oracle} & $6639$ & $6.973 \pm 1.534 $ & $1.00 \pm 0.00$ & $1.00 \pm 0.00$ & $0.720 \pm 0.044 $ & $0.151 \pm 0.022$ \\
                & $7519$ & $7.626 \pm 2.298 $ & $1.00 \pm 0.00$ & $1.00 \pm 0.00$ & $0.682 \pm 0.056 $ & $0.177 \pm 0.033$ \\
                & $1007$ & $7.895 \pm 1.106 $ & $1.00 \pm 0.00$ & $1.00 \pm 0.00$ & $0.726 \pm 0.045 $ & $0.207 \pm 0.025$ \\
                & $4520$ & $6.821 \pm 0.619 $ & $1.00 \pm 0.00$ & $1.00 \pm 0.00$ & $0.740 \pm 0.073 $ & $0.197 \pm 0.054$ \\
                & $4527$ & $9.431 \pm 3.057 $ & $1.00 \pm 0.00$ & $1.00 \pm 0.00$ & $0.700 \pm 0.056 $ & $0.174 \pm 0.017$ \\
\midrule
\textit{Worst case} & $6639$ & $595.3 \pm 297.9$ & $0.00 \pm 0.00$ & $1.00 \pm 0.00$ & $0.326 \pm 0.026 $ & $0.00 \pm 0.00$\\
                    & $7519$ & $317.2 \pm 235.8$ & $0.00 \pm 0.00$ & $1.00 \pm 0.00$ & $0.311 \pm 0.018 $ & $0.00 \pm 0.00$\\
                    & $1007$ & $508.1 \pm 155.7$ & $0.00 \pm 0.00$ & $1.00 \pm 0.00$ & $0.345 \pm 0.017 $ & $0.00 \pm 0.00$\\
                    & $4520$ & $299.6 \pm 124.2$ & $0.00 \pm 0.00$ & $1.00 \pm 0.00$ & $0.310 \pm 0.027 $ & $0.00 \pm 0.00$\\
                    & $4527$ & $432.8 \pm 72.05$ & $0.00 \pm 0.00$ & $1.00 \pm 0.00$ & $0.330 \pm 0.035 $ & $0.00 \pm 0.00$\\
                    
\bottomrule

\end{tabular}
\caption{Baseline results for Task $2$. Different structures. Averages and standard deviations reported based on five runs per seed.}
\label{tab:app_baseline_results_task_2}
\end{table*}

\section{Additional Results Task 1}
\label{sec:appendix_results_task1}

\begin{figure*}
     \centering
     \begin{subfigure}[b]{0.45\textwidth}
         \centering
         \includegraphics[width=\textwidth]{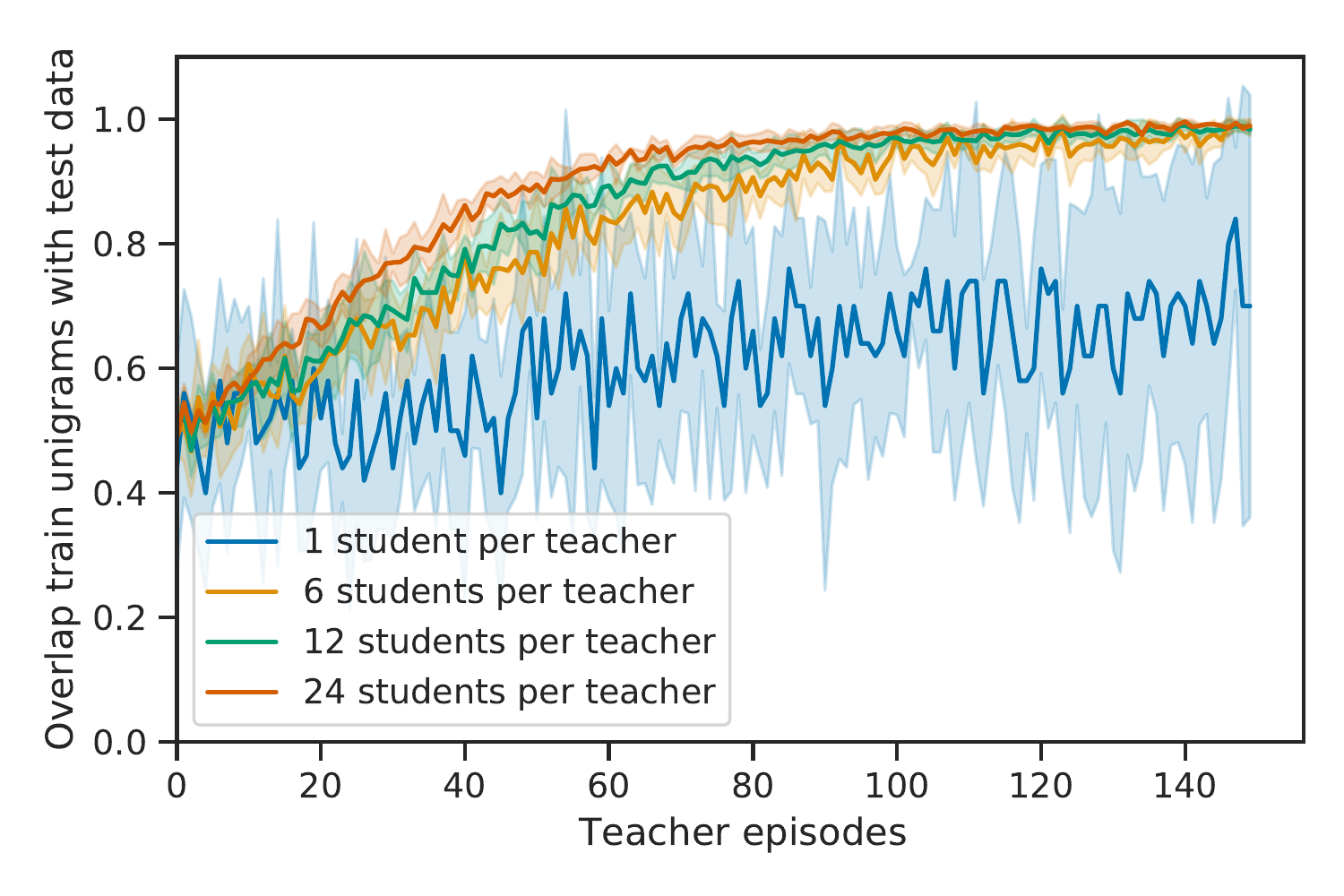}
         \vspace{-2em}
         \caption{Unigram overlap between train and test data.}
         \label{fig:task1_unigram_overlap_avg_word}
     \end{subfigure}
     \hfill
     \begin{subfigure}[b]{0.45\textwidth}
         \centering
         \includegraphics[width=\textwidth]{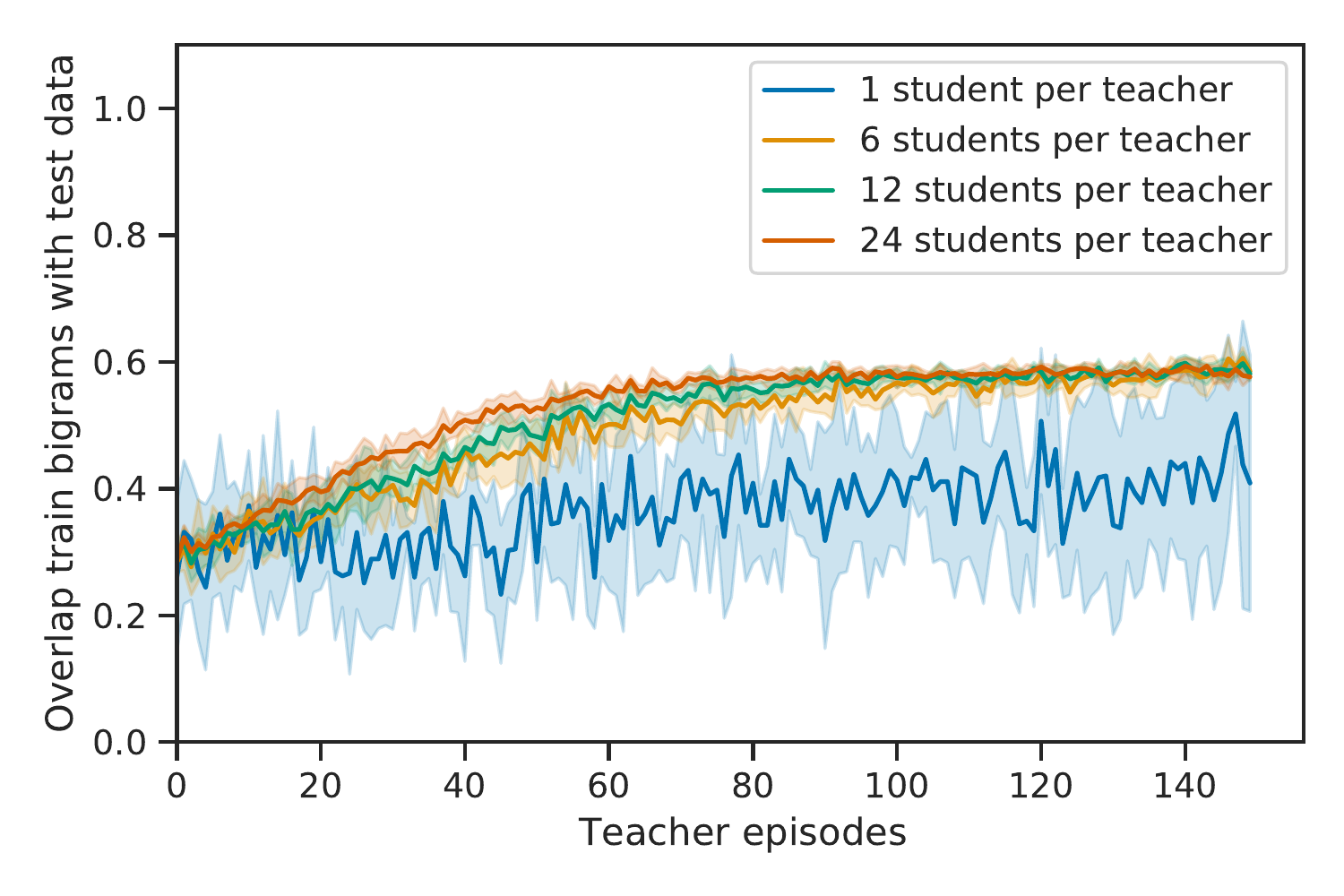}
         \vspace{-2em}
         \caption{Bigram overlap between train and test data.}
         \label{fig:task1_bigram_overlap_avg_word}
     \end{subfigure}
          \begin{subfigure}[b]{0.45\textwidth}
         \centering
         \includegraphics[width=\textwidth]{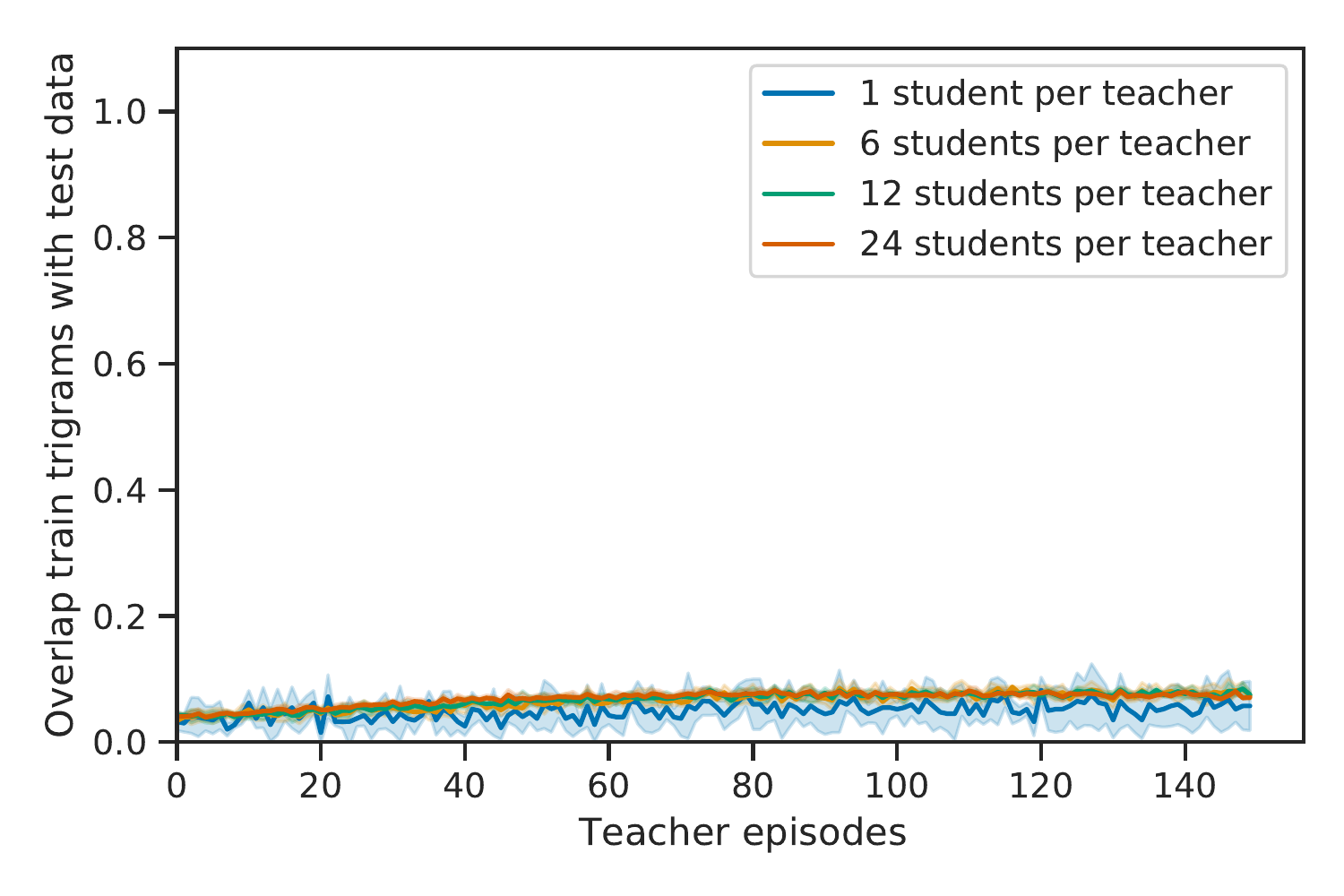}
         \vspace{-2em}
         \caption{Trigram overlap between train and test data.}
         \label{fig:task1_trigram_overlap_avg_word}
     \end{subfigure}
     \hfill
        \caption{Additional results Task 1 -- Different domains. Plots for different numbers of students per teacher. Results per setting reported as average and standard deviation over five random seeds. Average word embedding as sentence embeddings.}
        \vspace{-1em}
        \label{fig:appendix_results_task1_avg_word}
\end{figure*}

\begin{figure*}
     \centering
     \begin{subfigure}[b]{0.45\textwidth}
         \centering
         \includegraphics[width=\textwidth]{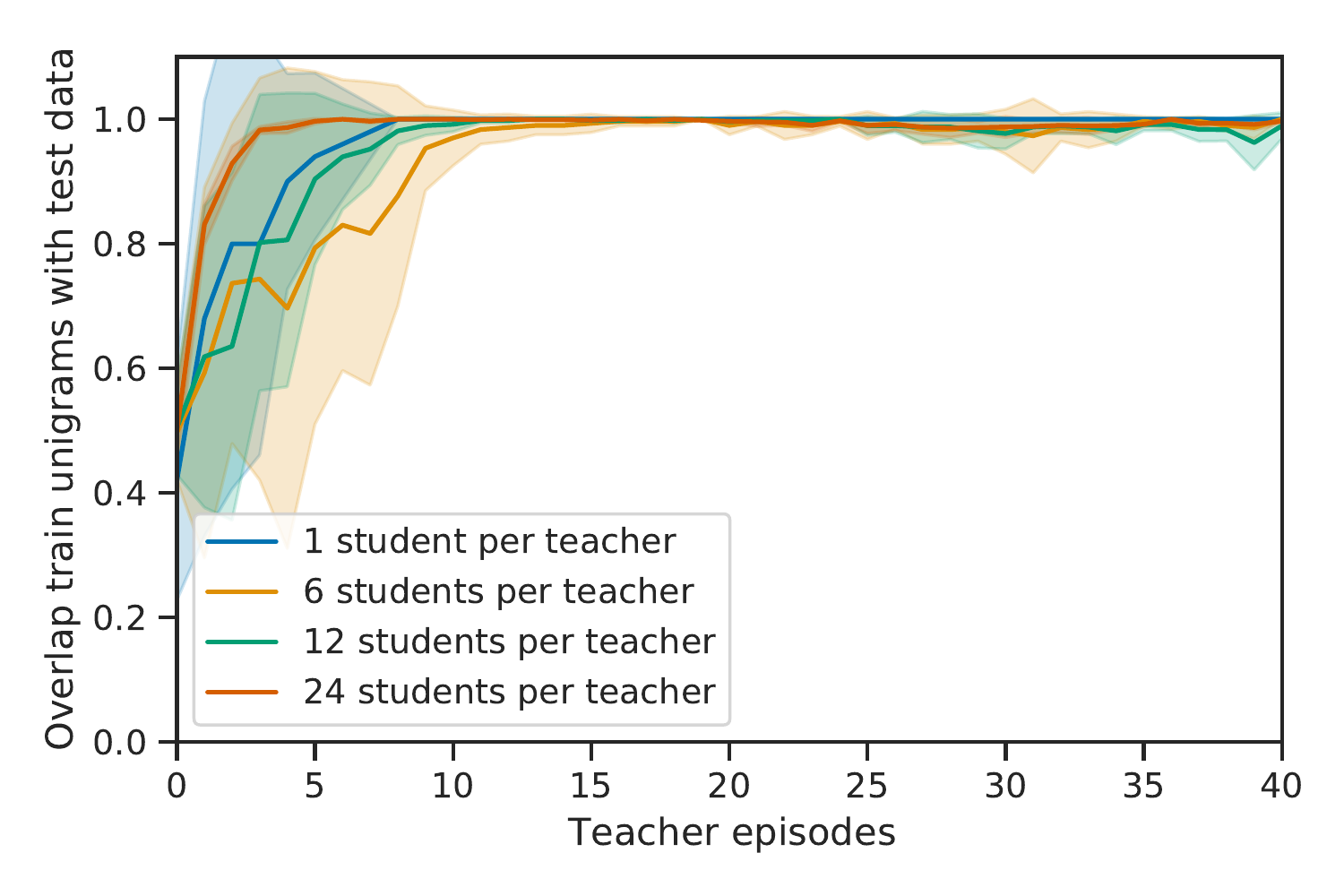}
         \vspace{-2em}
         \caption{Unigram overlap between train and test data.}
         \label{fig:task1_unigram_overlap_avg_hidden}
     \end{subfigure}
     \hfill
     \begin{subfigure}[b]{0.45\textwidth}
         \centering
         \includegraphics[width=\textwidth]{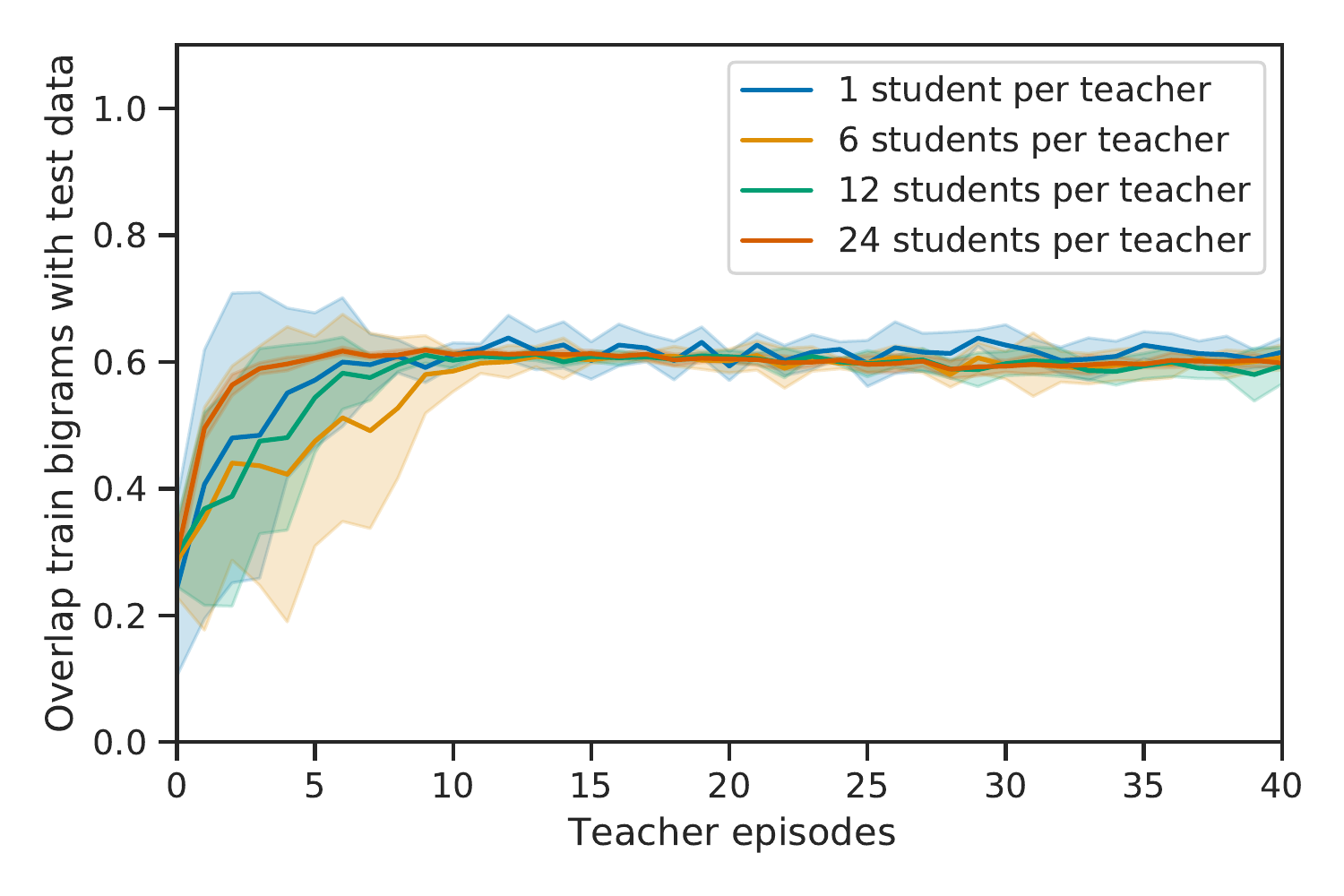}
         \vspace{-2em}
         \caption{Bigram overlap between train and test data.}
         \label{fig:task1_bigram_overlap_avg_hidden}
     \end{subfigure}
          \begin{subfigure}[b]{0.45\textwidth}
         \centering
         \includegraphics[width=\textwidth]{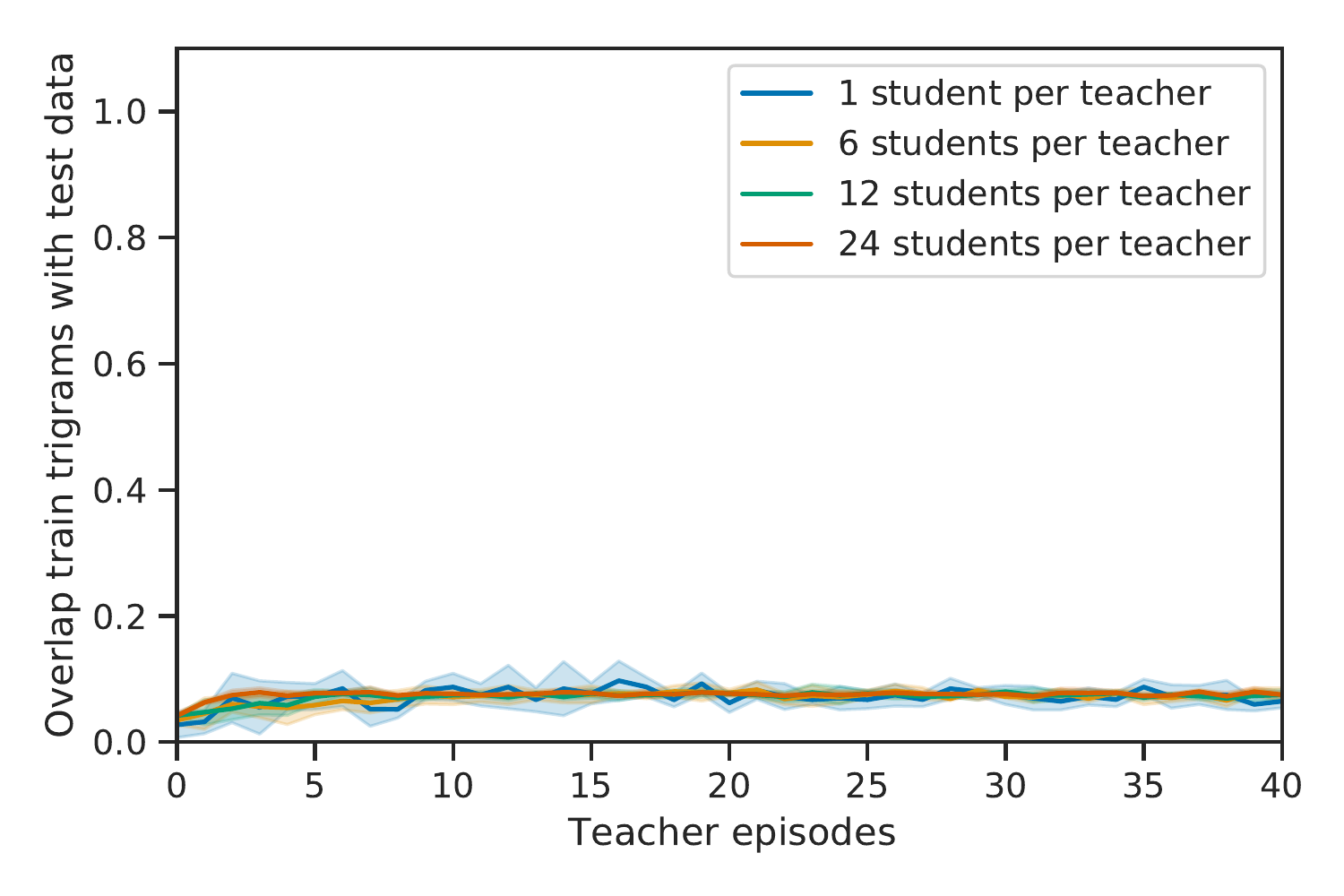}
         \vspace{-2em}
         \caption{Trigram overlap between train and test data.}
         \label{fig:task1_trigram_overlap_avg_hidden}
     \end{subfigure}
     \hfill
        \caption{Additional results Task 1 -- Different domains. Plots for different numbers of students per teacher. Results per setting reported as average and standard deviation over five random seeds. Average hidden layer embedding as sentence embeddings.}
        \vspace{-1em}
        \label{fig:appendix_results_task1_avg_hidden}
\end{figure*}

In this section we present the plots for the $n$-gram overlap for Task $1$ in Figures~\ref{fig:appendix_results_task1_avg_word} and~\ref{fig:appendix_results_task1_avg_hidden}.

\section{Additional Results Task 2}
\label{sec:appendix_results_task2}

\begin{figure*}[h]
     \centering
     \begin{subfigure}[b]{0.45\textwidth}
         \centering
         \includegraphics[width=\textwidth]{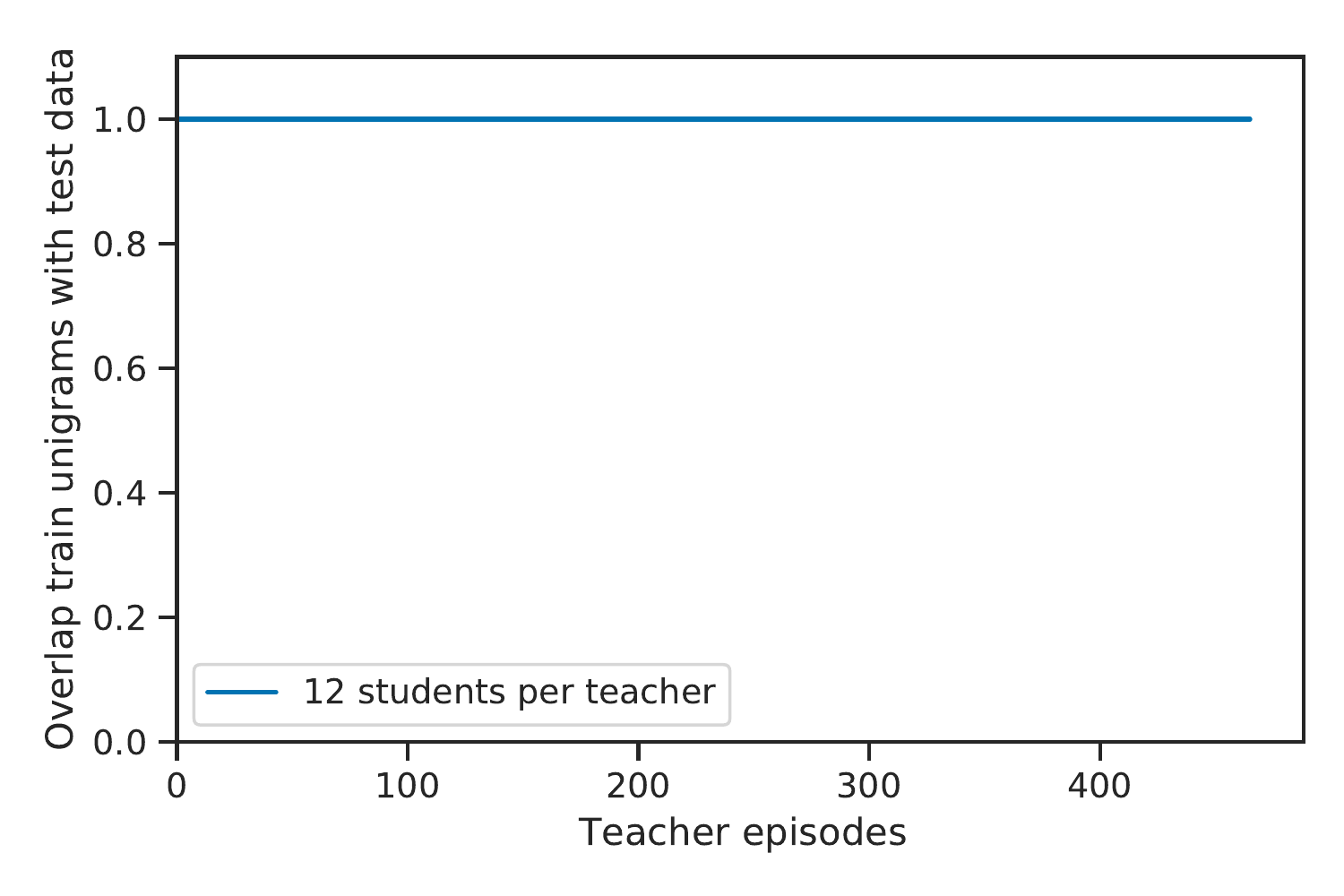}
         \vspace{-2em}
         \caption{Unigram overlap between train and test data.}
         \label{fig:task2_unigram_overlap}
     \end{subfigure}
     \hfill
     \begin{subfigure}[b]{0.45\textwidth}
         \centering
         \includegraphics[width=\textwidth]{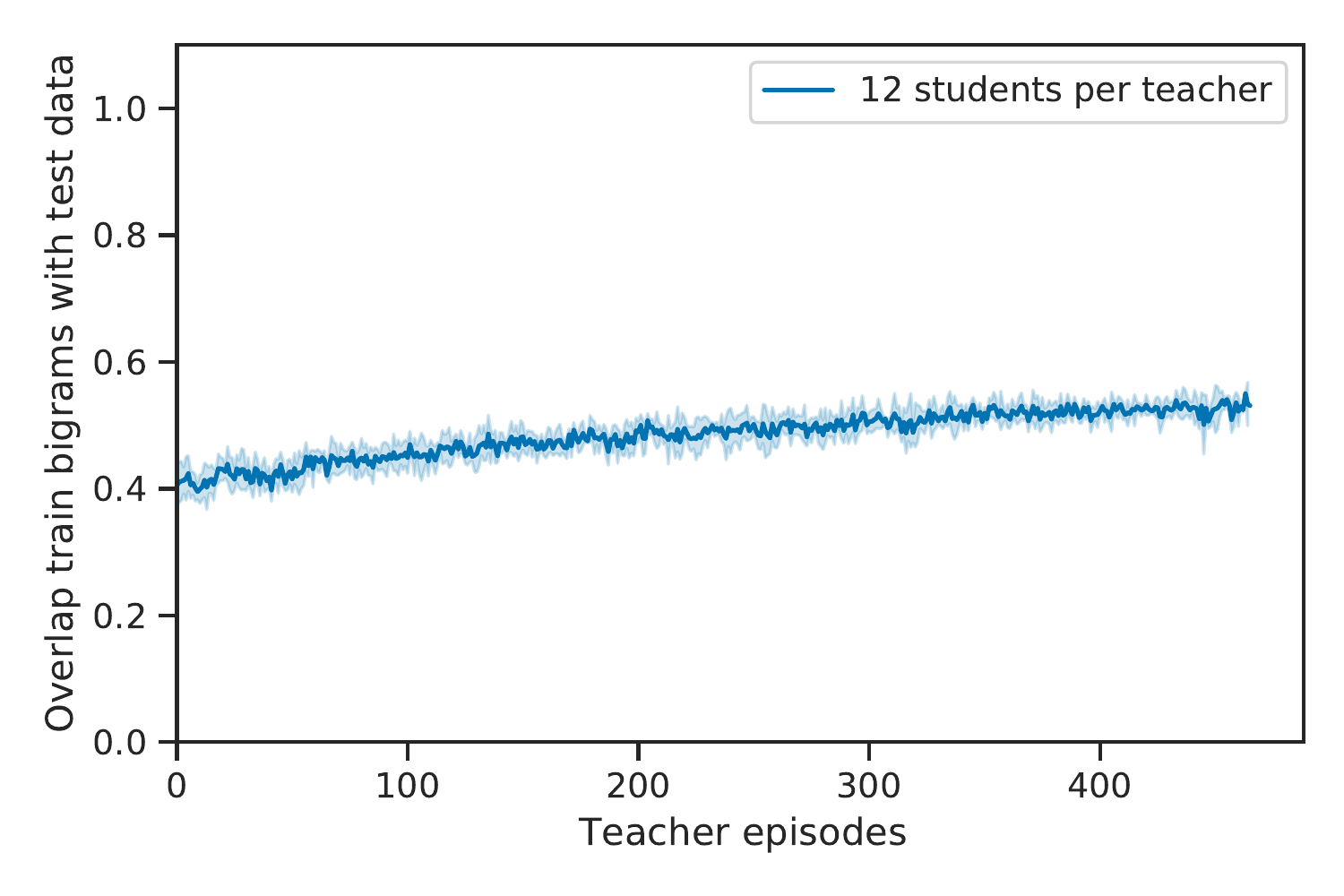}
         \vspace{-2em}
         \caption{Bigram overlap between train and test data.}
         \label{fig:task2_bigram_overlap}
     \end{subfigure}
          \begin{subfigure}[b]{0.45\textwidth}
         \centering
         \includegraphics[width=\textwidth]{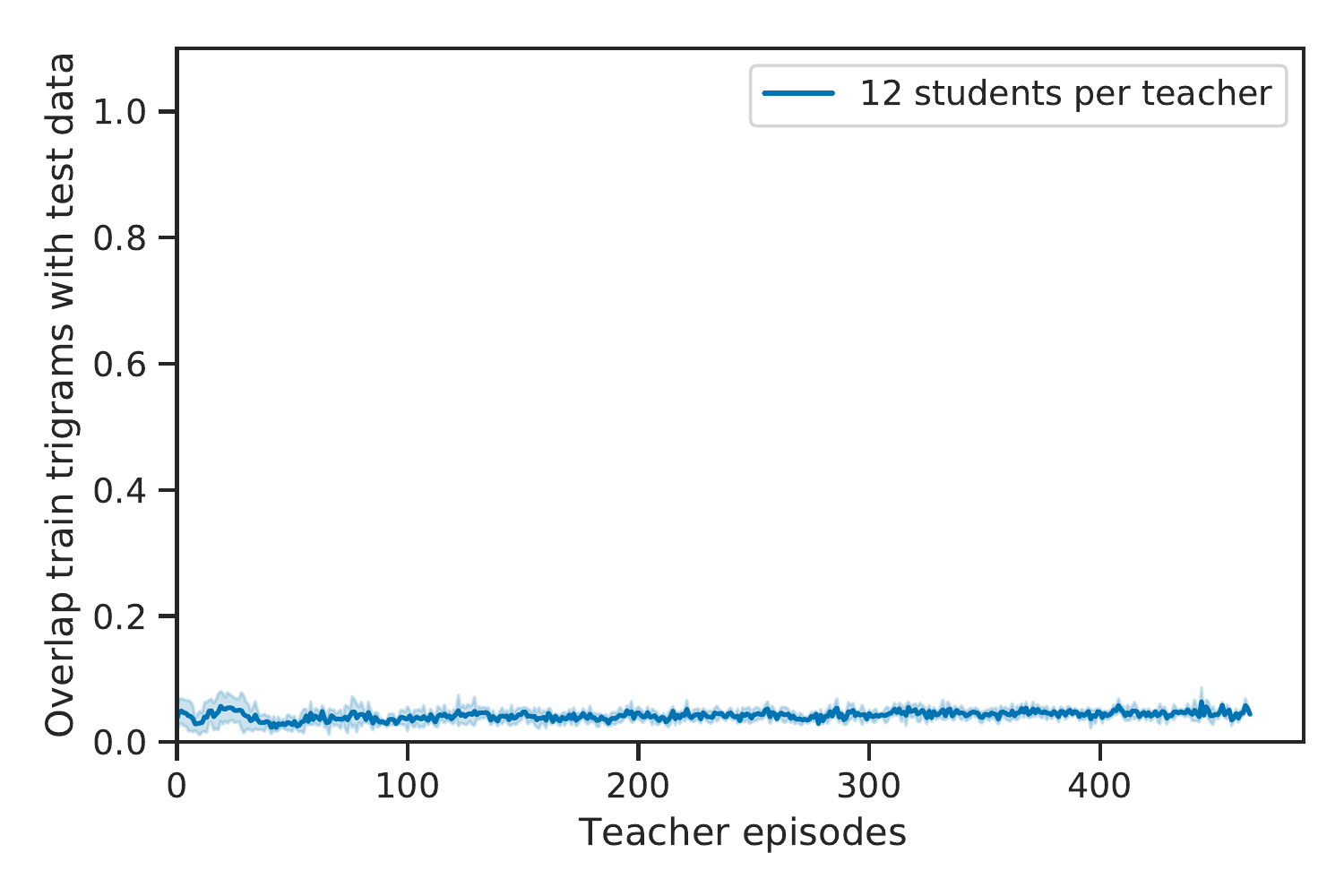}
         \vspace{-2em}
         \caption{Trigram overlap between train and test data.}
         \label{fig:task2_trigram_overlap}
     \end{subfigure}
     \hfill
        \caption{Additional results Task 2 -- Different structures. Results per setting reported as average and standard deviation over five random seeds.}
        \vspace{-1em}
        \label{fig:appendix_results_task2}
\end{figure*}

In this section we present the plots for the $n$-gram overlap for Task $2$ in Figure~\ref{fig:appendix_results_task2}.

